\documentclass[lettersize,journal]{IEEEtran}
\usepackage{amsmath,amsfonts}
\usepackage{algorithmic}
\usepackage{algorithm}
\usepackage{array}
\usepackage{textcomp}
\usepackage{subfigure}
\usepackage{stfloats}
\usepackage{url}
\usepackage{verbatim}
\usepackage{graphicx}
\usepackage{cite}

\usepackage{multirow}
\usepackage{booktabs}
\usepackage{colortbl}  
\usepackage{xcolor}

\begin{document}

\title{Zoom Text Detector}

\author{
	Chuang~Yang,
	Mulin~Chen,
	Yuan~Yuan,~\IEEEmembership{Senior Member,~IEEE,}
	and~Qi~Wang,~\IEEEmembership{Senior Member,~IEEE}
	
	\thanks{
		Chuang~Yang is with the School of Computer Science, and with the School of Artificial Intelligence, OPtics and ElectroNics (iOPEN), Northwestern Polytechnical University, Xi'an 710072, Shaanxi, P. R. China. Mulin~Chen, Yuan~Yuan, and Qi~Wang are with the School of Artificial Intelligence, OPtics and Electronics (iOPEN), Northwestern Polytechnical University, Xi'an 710072, P.R. China. E-mail: cyang113@mail.nwpu.edu.cn, chenmulin@mail.nwpu.edu.cn, y.yuan.ieee@gmail.com, crabwq@gmail.com.
			}
	
	\thanks{Qi Wang is the corresponding author.
			}
	   }



\maketitle

\begin{abstract}
To pursue comprehensive performance, recent text detectors improve detection speed at the expense of accuracy. They adopt shrink-mask based text representation strategies, which leads to a high dependency of detection accuracy on shrink-masks. Unfortunately, three disadvantages cause unreliable shrink-masks. Specifically, these methods try to strengthen the discrimination of shrink-masks from the background by semantic information. However, the feature defocusing phenomenon that coarse layers are optimized by fine-grained objectives limits the extraction of semantic features. Meanwhile, since both shrink-masks and the margins belong to texts, the detail loss phenomenon that the margins are ignored hinders the distinguishment of shrink-masks from the margins, which causes ambiguous shrink-mask edges. Moreover, false-positive samples enjoy similar visual features with shrink-masks. They aggravate the decline of shrink-masks recognition. To avoid the above problems, we propose a Zoom Text Detector (ZTD) inspired by the zoom process of the camera. Specifically, Zoom Out Module (ZOM) is introduced to provide coarse-grained optimization objectives for coarse layers to avoid feature defocusing. Meanwhile, Zoom In Module (ZIM) is presented to enhance the margins recognition to prevent detail loss. Furthermore, Sequential-Visual Discriminator (SVD) is designed to suppress false-positive samples by sequential and visual features. Experiments verify the superior comprehensive performance of ZTD.
\end{abstract}


\section{Introduction}
\label{introduction}
\IEEEPARstart{T}{ext} detection, the key to retrieving texts, has become an attractive topic and involves various applications (such as multilingual translation systems and unmanned systems). In the past decade, since deep learning technologies~\cite{long2015fully,redmon2016you,tian2019fcos} have shown impressive performance in computer vision and artificial intelligence, many deep learning-based algorithms are proposed for text detection~\cite{zhang2022kernel}, which can be categorized into two classes roughly: accuracy prior methods~\cite{DBLP:conf/cvpr/FengYZL21,zhang2020opmp} and comprehensive performance prior methods~\cite{zhou2017east,wang2019efficient,liao2020real}. 

The former represents text instances by multiple local units or rebuilds text contours by a series of geometry operations. These methods usually enjoy high detection accuracy. However, the complicated frameworks lead to expensive memory overhead, deep dependency for high-performance computing units, and slow inference. Furthermore, related works~\cite{zhu2021fourier,ma2021relatext,8812908} show the weak gains for model accuracy with the increase of model complexity. The latter aims to accelerate the inference process with lightweight frameworks to make it possible to deploy text detection techniques into mobile terminals. These works~\cite{wang2019efficient,liao2020real} model the whole text instances directly through shrink-mask based text representation strategies. They only need to conduct prediction tasks on one feature map that is fused by multi-level feature maps and rebuild text contours by simple post-processing, which simplifies the frameworks and improves the detection speed effectively. However, three disadvantages exist in these methods, which limit the improvement of detection accuracy. 

\begin{figure}
	\begin{center}
		\includegraphics[width=0.45\textwidth]{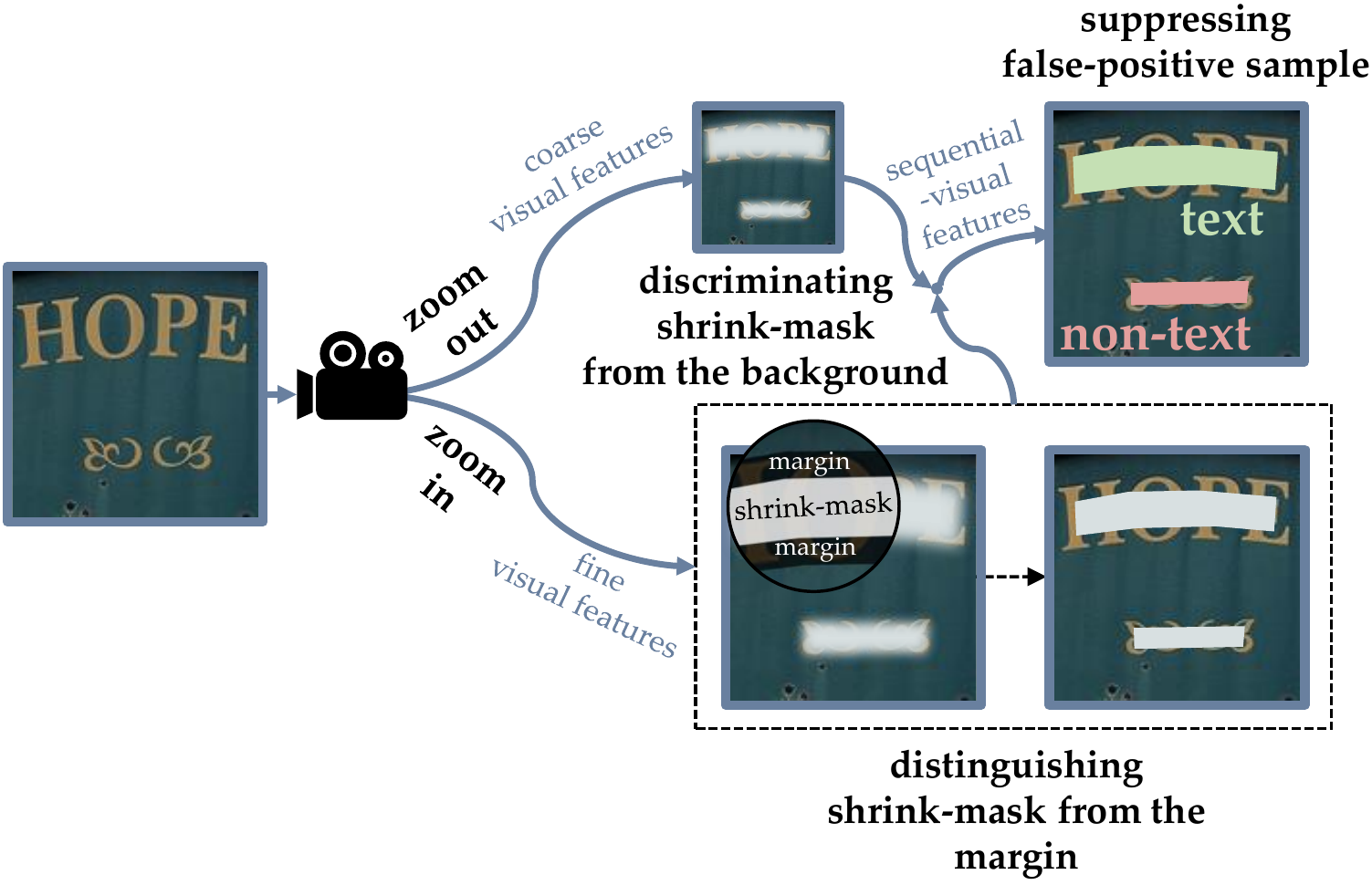}
	\end{center}
	\vspace{-5mm}
	\caption{Motivation of the proposed method. We aim to avoid feature defocusing and detail loss by simulating the zoom process of the camera. Meanwhile, we suppress false-positive samples according to the combination of sequential and visual features.}
	\label{V1}
	\vspace{-5mm}
\end{figure}

The first one is the phenomenon of feature defocusing. Shrink-mask based text representation strategies lead to the detection accuracy being highly dependent on shrink-masks. To pursue reliable shrink-masks, current algorithms merge coarse layers into fine layers following the idea of~\cite{long2015fully}. They try to enhance the discrimination of shrink-masks from the background by the semantic information from coarse layers. However, the layers are only supervised by the fine-grained optimization objectives in the training stage, which limits the extraction of semantic features. The second one is the phenomenon of detail loss. It is found in the Contour Extension Process in Fig.~\ref{V3}, both shrink-masks and the corresponding margins belong to texts, which makes it hard to determine clear borders between shrink-masks and the margins. Existing methods ignore the margins recognition. It accelerates the decline of the model's ability to distinguish shrink-masks from the margins, which results in ambiguous shrink-mask edges and leads to many adverse effects (such as text adhesion, miss detection, and the introduction of noise information). Moreover, false-positive samples enjoy highly similar visual features (such as color, texture, geometry, etc.) with shrink-masks. However, previous detectors suppress them according to visual features only, which aggravates the decline of model accuracy.

Considering the limitations above, how to overcome those problems is still under explored. The photographers capture global information of scenes by zooming out the camera, which helps to analyze the semantic relationships between different objects. Meanwhile, they zoom-in the camera to focus on local regions, which supports observing object details. In this paper, inspired by the zoom process of the camera, we propose Zoom Text Detector (ZTD). It makes full use of the advantages of coarse and fine features to enhance the reliability of shrink-masks. Specifically, as shown in Fig.~\ref{V1}, to help discriminate shrink-masks from the background, a Zoom Out Module (ZOM) simulates the zoom-out process of the camera to focus on coarse features. It provides coarse-grained optimization objectives for coarse layers to facilitate the extraction of coarse features with strong semantic information. Meanwhile, to strengthen the distinguishment of shrink-masks from the corresponding margins, a Zoom In Module (ZIM) simulates the zoom-in process of the camera to utilize fine features to to enhance ZTD's ability to recognize the margins. Moreover, considering shrink-masks are equipped with rich sequential features, a Sequential-Visual Discriminator (SVD) is designed to combine sequential and visual features to help suppress false-positive samples. The main contributions of this paper are as follows:

\begin{enumerate}
	\item Zoom strategy-based Zoom Out Module (ZOM) and Zoom In Module (ZIM) are proposed, which maximize the advantages of coarse and fine features to avoid feature defocusing and detail loss. The former helps to discriminate shrink-masks from the background. The latter strengthens the distinguishment of shrink-masks from the margins.
	
	\item A Sequential-Visual Discriminator (SVD) is designed to encourage ZTD to learn the combination of sequential and visual features, which helps to suppress false-positive samples from temporal and spatial domains. Particularly, it brings no computational cost to the inference process and can be integrated into other detectors seamlessly.
	
	\item An efficient text detection framework combined with a lightweight CNN model and simple post-processing is constructed. It achieves the detection accuracy comparable with accuracy prior methods and runs faster than comprehensive performance prior algorithms, which provides sufficient support for practical applications.
\end{enumerate}

The rest of the paper is organized as follows. Section~\ref{sec2} introduces the related works on text detection. Section~\ref{sec3} describes the structure of ZTD. The experimental results are discussed in Section~\ref{sec4}. Section~\ref{sec5} concludes the paper.

\section{Related Work}
\label{sec2}
In recent years, deep learning-based methods have achieved dominant performance on text detection, which can be classified into accuracy prior methods and comprehensive performance prior methods roughly.

\subsection{Accuracy Prior Methods}
Two-stage detection methods, such as Faster-RCNN~\cite{ren2015faster}, have brought great inspiration to text detection~\cite{liu2017deep,ma2018arbitrary,zhang2019look} early. To speed up the text inference process, more one-stage methods~\cite{shi2017detecting,tian2017wetext} based on SSD~\cite{liu2016ssd} were proposed. For extracting text features with strong expression, Liao~$et~al$.~\cite{liao2016textboxes} proposed 1$\times$5 convolution kernel to fit text geometries, and He~$et~al$.~\cite{he2017single} introduced attention mechanism into the text detection model. To detect multi-oriented texts, several extra works~\cite{liao2018textboxes++,ma2018arbitrary,wang2021towards}  predicted angles of multi-oriented texts. Others~\cite{liao2018rotation,xu2020gliding} regressed the offsets between bounding boxes and anchors. Since anchor mechanism increases the model complexity largely, anchor-free text detector~\cite{wang2020textray} based on ~\cite{huang2015densebox,li2017multiview,tian2019fcos} has been proposed, which obtained multiple contour points through regression strategy directly. Recently, researchers focus on the representation of irregular-shaped text contours. Some 
works~\cite{DBLP:conf/aaai/DengLLC18,DBLP:conf/cvpr/ZhangZHLYWY20,ma2021relatext} detected multiple character-level bounding boxes by regression strategy and linked the boxes to obtain the final text contours. Different from them, other works~\cite{long2018textsnake,baek2019character} predicted character-level boxes based on segmentation technology. Furthermore, the authors~\cite{law2018cornernet,wang2020contournet} and~\cite{wang2020all,wang2020textray,xie2021polarmask++} modeled text contours as a series of dense contour key points. They only needed to obtain the points through segmentation and regression technologies, respectively. In addition, latest works~\cite{zhu2021fourier} introduced mathematical model to fit text instances. Specifically Zhu~$et~al$.~\cite{zhu2021fourier} modeled text instances in the Fourier domain and expressed arbitrary-shaped text contours as compact signatures. Except for the above algorithms, Tian~$et~al$.~\cite{tian2019learning} and Xu~$et~al$.~\cite{xu2019textfield} segmented the whole text regions directly and separated adjacent texts by direction maps. Though these methods achieve high detection accuracy for arbitrary-shaped text detection, the complicated framework hinders the feasibility in practical applications.

\subsection{Comprehensive Performance Prior Methods}
To meet the high requirements of the practical application scenarios for the comprehensive performance of algorithms, some approaches~\cite{zhou2017east,wang2019efficient,liao2020real} improved detection speed at the expense of accuracy by simplifying the framework. Inspired by ~\cite{huang2015densebox}, Zhou~$et~al$.~\cite{zhou2017east} presented a one-stage text detection method, which optimized the inference process largely after abandoning anchor mechanism. Moreover, the authors augmented positive samples in the training stage by introducing the shrink-mask. Recently, to further simplify the framework while handling irregular-shaped texts, most works are dedicated to researching an efficient text representation model. Wang~$et~al$.~\cite{wang2019efficient,wang2021pan++} proposed a faster region extension strategy based on their previous work~\cite{wang2019shape}, which can effectively fit arbitrary-shaped text instances and overcome the text adhesion problem. Although the methods enjoy high detection speed, pixel-wise extension based post-processing is relatively time-consuming. Considering this problem, Liao~$et~al$.~\cite{liao2020real} designed an object-wise extension strategy based text detection framework, which only needed to predict shrink-masks by one segmentation header. Importantly, the object-wise strategy saved much computational cost compared with~\cite{wang2019efficient}. 
\begin{figure*}
	\begin{center}
		\includegraphics[width=0.9\textwidth]{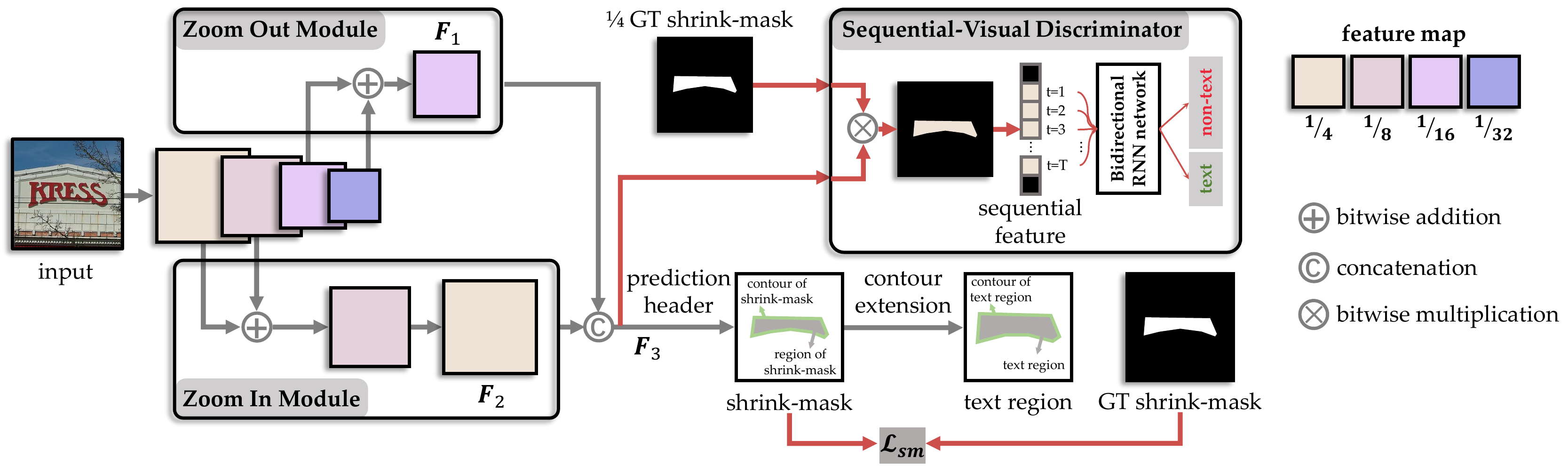}
	\end{center}
	\vspace{-5mm}
	\caption{Overall structure of the proposed Zoom Text Detector, which consists of Zoom Out Module, Zoom In Module, Sequential-Visual Discriminator, shrink-mask prediction header, and contour extension process. Red flows are training only operators.}
	\label{V2}
\end{figure*}
\begin{figure}
	\begin{center}
		\includegraphics[width=0.45\textwidth]{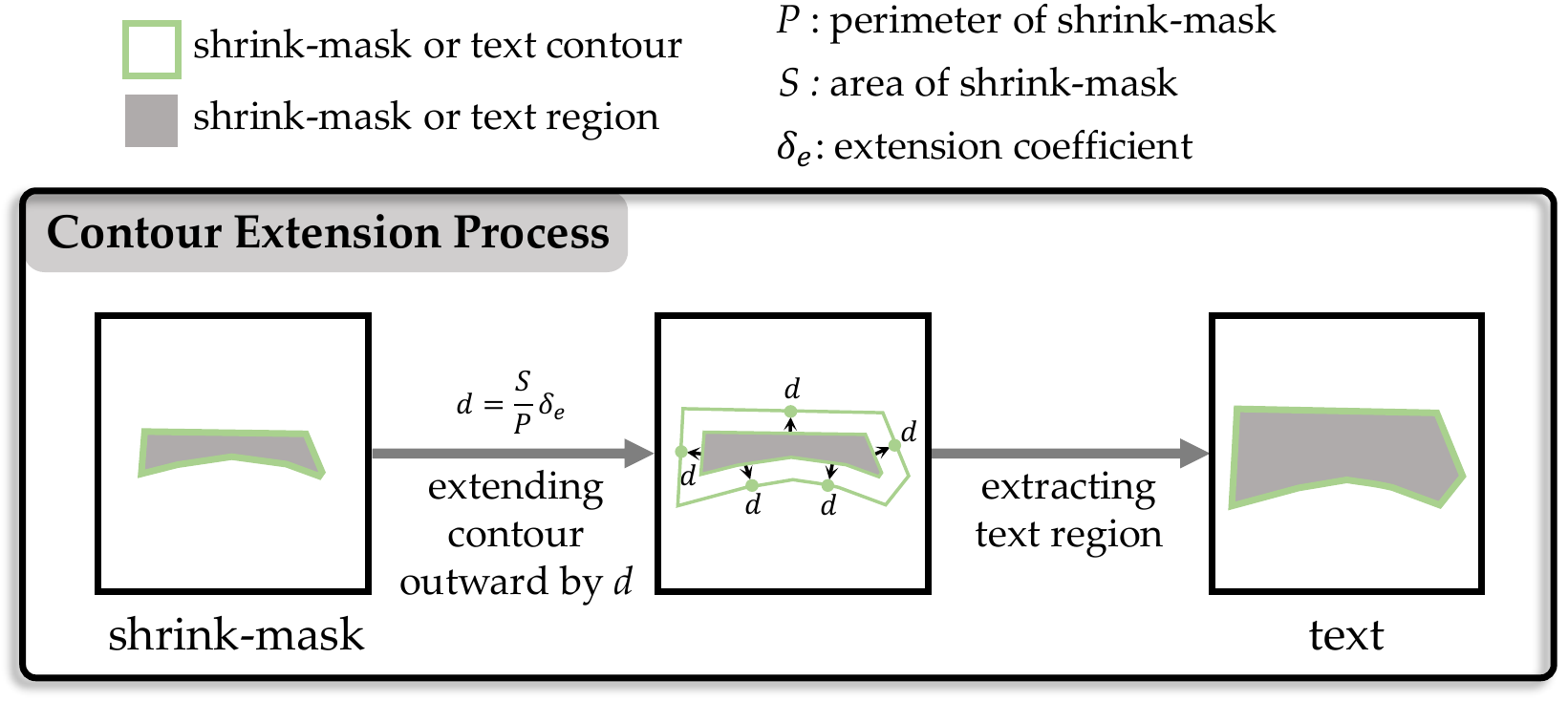}
	\end{center}
	\vspace{-5mm}
	\caption{Visualization of text contour extension process.}
	\label{V3}
	\vspace{-2mm}
\end{figure}

\section{Methodology}
\label{sec3}
In this section, we introduce the overall structure of the proposed ZTD firstly. Then, the details of Zoom Out Module (ZOM), Zoom In Module (ZIM), and Sequential-Visual Discriminator (SVD) are described and shown through visualization. In the end, the optimization function is given.

\subsection{Overall Pipeline}
\label{overall}
The architecture of ZTD is shown in Figure~\ref{V2}, which is composed of backbone, Zoom Out Module (ZOM), Zoom In Module (ZIM), Sequential-Visual Discriminator (SVD), prediction header, and contour extension process. For backbone, it is used for the generation of multi-level feature maps $f_1$, $f_2$, $f_3$, and $f_4$ corresponding to the image size of $\frac{1}{4}$, $\frac{1}{8}$, $\frac{1}{16}$, and $\frac{1}{32}$, respectively. To avoid the phenomena of feature defocusing and detail loss that exist in current methods~\cite{wang2019efficient,liao2020real}, ZOM and ZIM are proposed. Specifically, as shown in Fig.~\ref{V2}, ZOM fuses $f_3$ and $f_4$ at first. Then, a coarse segmentation task is conducted on $F_1$ to help ZTD to extract semantic features, which promotes the discrimination of shrink-masks from the background. For ZIM, it utilizes fine features to force ZTD to recognize the margins, which facilitates our method to distinguish shrink-masks from the margins. Considering false-positive samples enjoy similar visual features (such as color, texture, and edge) with shrink-masks, SVD encourages ZTD to extract the sequential and visual features to distinguish them in the temporal and spatial domains. The header consists of two transposed convolution layers. It utilizes the hybrid features from ZOM, ZIM, and SVD to predict shrink-masks accurately. In the end, the texts can be rebuilt by the contour extension process (as shown in Fig.~\ref{V3}) directly, where the extension distance is computed by the formula in~\cite{vatti1992generic}.

\begin{figure}
	\begin{center}
		\includegraphics[width=0.45\textwidth]{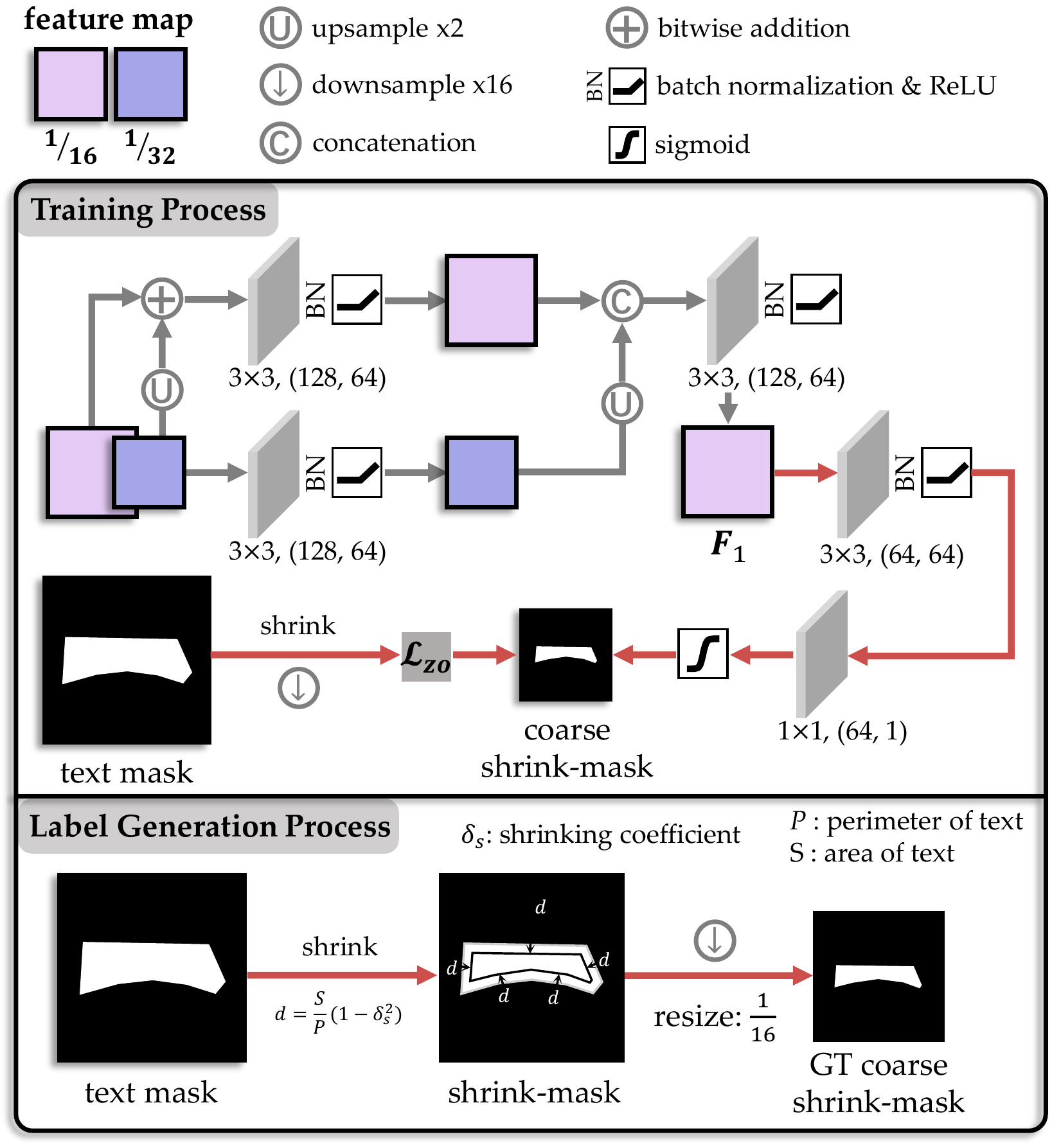}
	\end{center}
	\vspace{-5mm}
	\caption{Visualization of the structure and label generation process of Zoom Out Module. Red flows are training only operators}
	\label{V4}
	\vspace{-2mm}
\end{figure}

\begin{figure*}
	\begin{center}
		\includegraphics[width=0.95\textwidth]{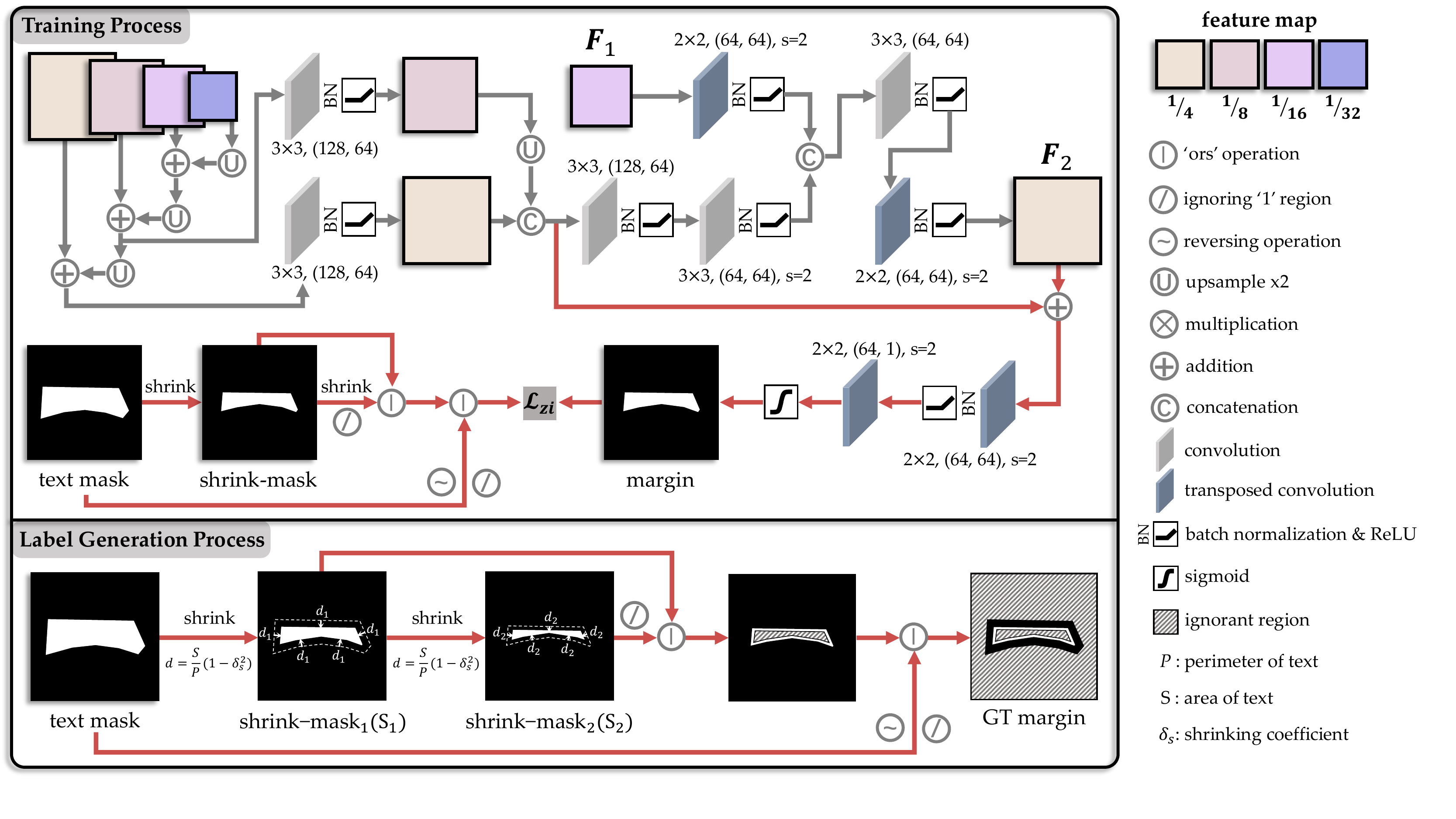}
	\end{center}
	\vspace{-5mm}
	\caption{Visualization of the structure and label generation process of Zoom In Module. Red flows are training only operators}
	\label{V5}
	\vspace{-2mm}
\end{figure*}

\subsection{Zoom Out Module}
According to~\cite{long2015fully}, semantic information can strengthen discrimination of shrink-masks from the background. However, for recent lightweight methods~\cite{wang2019efficient,liao2020real}, the phenomenon of feature defocusing limits the extraction of semantic features, where coarse layers are optimized by the fine-grained supervision information. Considering the above problems, ZOM is proposed to provide coarse-grained optimization objectives for coarse layers to facilitate the extraction of semantic features.

The structure of ZOM is shown in Fig.~\ref{V4}. The module fuses two coarse features to strengthen the expression of $f_4$ firstly. To save computational cost, we use two $3\times3$ convolution layers to reduce the channels of feature maps in $f_3$ and $f_4$ branches, respectively. Considering the concatenation operator can provide a larger mapping space compared to point-wise addition and experimental results in~\cite{ronneberger2015u} verify the effectiveness of the concatenation layer for segmentation tasks, $f_3$ and $f_4$ are concatenated in this structure. Following the design strategy of the lightweight network, we reduce the concatenated feature channels to half of the original to generate the feature map $F_1$.

In the training process, a $3\times3$ convolution is used for $F_1$ smoothing, and a $1\times1$ convolution is conducted on the smooth features to segment the coarse shrink-mask map. The shrink-mask can be obtained by performing a sigmoid function on the map. The loss function $L_{zo}$ evaluates the error between the predicted coarse shrink-mask and the corresponding ground-truth. Furthermore, we visualize the label generation process of the coarse shrink-mask in Fig.~\ref{V4}. Specifically, text contour is shrunk inward by $d$ that is computed by the formula in~\cite{vatti1992generic} and the region in shrunk contour is treated as shrink-mask. The ground-truth of coarse shrink-mask is obtained by resizing shrink-mask to $\frac{1}{16}$ of image size. Particularly, except the final $1\times1$ convolution, each convolutional layer is followed by a BatchNorm layer~\cite{ioffe2015batch} and rectified linear unit (ReLU)~\cite{glorot2011deep}. 

\subsection{Zoom In Module}
The shrink-mask is generated by shrinking the text contour (as shown in Fig.~\ref{V4}). It means both shrink-masks and the corresponding margins belong to texts, which makes it hard to discriminate them. For current methods, they ignore the margins recognition (the phenomenon of detail loss), which leads to ambiguous shrink-mask edges and further results in text adhesion and miss detection. Considering fine-grained information is useful for discriminating shrink-masks and the margins, ZIM is designed to encourage ZTD to utilize the information to recognize the margins, which helps to predict shrink-mask edges precisely. 

The architecture of ZIM is illustrated in Fig.~\ref{V5}. In the front part, it conducts the same operations on $f_1$ and $f_2$ like ZOM to $f_3$ and $f_4$. The concatenated feature is treated as an input for the following two branches. For the first one, the feature is downsampled by a $3\times3$ convolution with two strides after reducing channels. Then, it is concatenated with upsampled $F_1$ from ZOM, where $F_1$ is used to provide semantic information for enhancing fine-grained margin recognition. Next, we further upsample the feature by transposed convolution as $\frac{1}{4}$ size of image for pixel-wise segmentation. In the end, considering the image details are lost with the increase of network layers, the above concatenated feature is skip connected with the upsampled feature $F_2$.  

In the training process, the combination of two transposed convolutions and one sigmoid function is used to predict the margins, which is helpful to strengthen the model's ability to recognize shrink-mask edges. The loss function $L_{zi}$ evaluates the differences between the predicted margins and the corresponding ground-truth. As shown in Fig.~\ref{V5}, the label generation process includes five steps: 1) Shrinking text contour to obtain the shrink-mask $S_1$; 2) Shrinking the contour of $S_1$ to obtain a smaller shrink-mask $S_2$; 3) Ignoring the '1' region in $S_2$ and conducting 'ors' operation between $S_1$ and ignored $S_2$; 4) Reversing text mask and ignoring the '1' region; 5) Conducting 'ors' operation between the processed text mask and the result generated by step 3. The mentioned 'ors' operation is defined as:
\begin{eqnarray}
\begin{gathered}
p_{i,j}^{a}~{\rm ors}~p_{i,j}^{b}=ignorance,\\
\{p_{i,j}^{a}=ignorance~{\rm or}~p_{i,j}^{b}=ignorance\},
\end{gathered}
\end{eqnarray}
\begin{eqnarray}
\begin{gathered}
p_{i,j}^{a}~{\rm ors}~p_{i,j}^{b}=1,\\
\{p_{i,j}^{a}=1~{\rm or}~p_{i,j}^{b}=1\},\\
\{p_{i,j}^{a}\neq ignorance~{\rm and}~p_{i,j}^{b}\neq ignorance\},
\end{gathered}
\end{eqnarray}
where $p_{i,j}^{a}$ and $p_{i,j}^{b}$ denote the pixel category of $i$th row and $j$th column on mask $a$ and $b$, respectively.

\subsection{Sequential-Visual Discriminator}
False-positive samples enjoy highly similar visual features with shrink-masks (such as color, texture, and edge), which makes it difficult to discriminate them according to visual features only. Considering shrink-masks are equipped with sequential features, SVD is designed to encourage ZTD to utilize the combination of sequential and visual features to suppress false-positive samples.

\begin{figure}
	\begin{center}
		\includegraphics[width=0.45\textwidth]{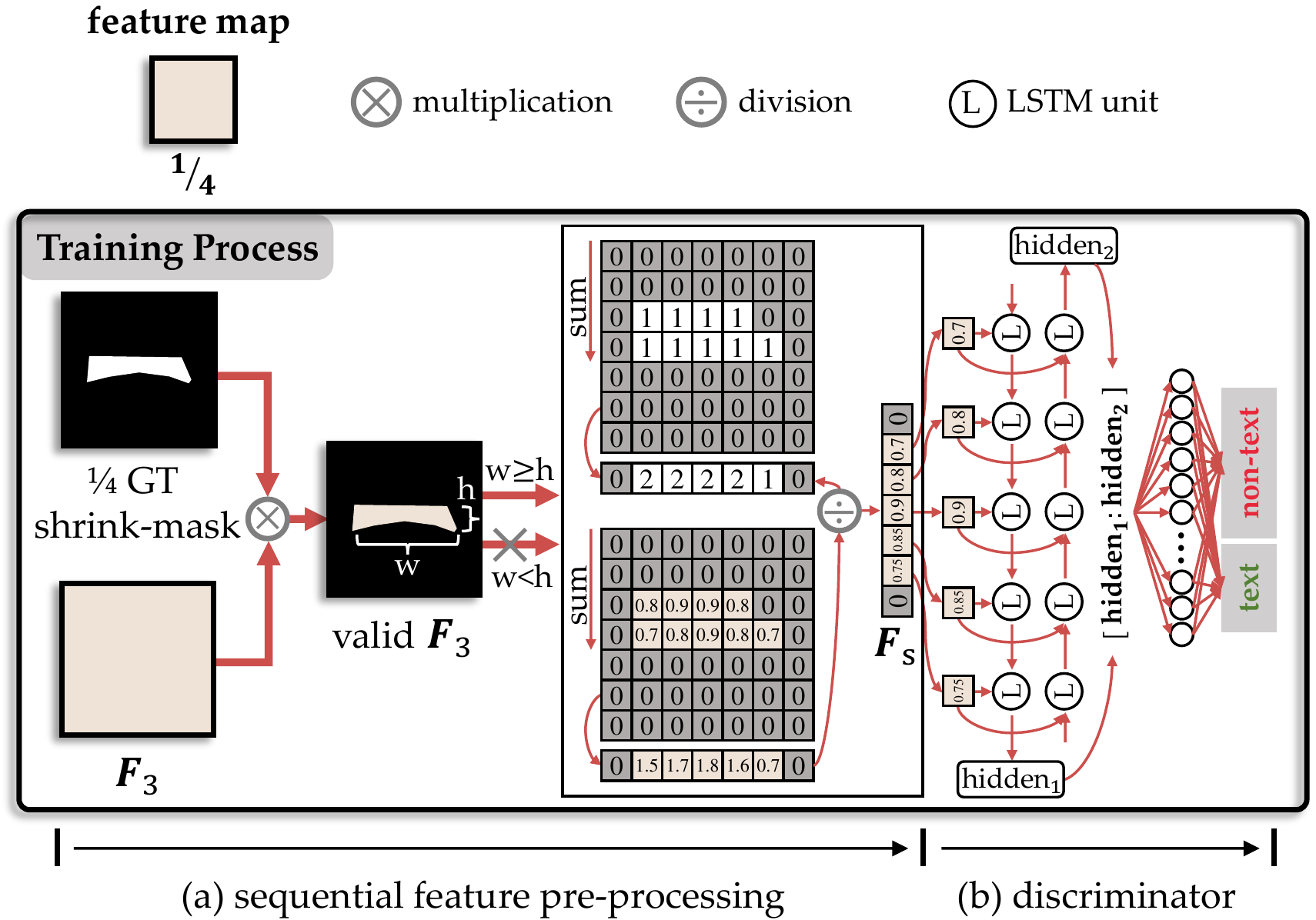}
	\end{center}
	\vspace{-3mm}
	\caption{Visualization of the structure of Sequential-Visual Discriminator. Red flows are training only operators.}
	\label{V6}
\end{figure}

Details of SVD are shown in Fig.~\ref{V6}. The module consists of sequential feature pre-processing and discriminator. The pre-processing takes $\frac{1}{4}$ GT shrink-mask and ${\rm F_3}$ (shown in Fig.~\ref{V2}) as input. It first executes multiplication on them to generate valid ${\rm F_3}$ and defines non-zero region as valid feature region. Then, the width $w$ and height $h$ of the region are computed, and the pixels of $\frac{1}{4}$ GT shrink-mask and valid ${\rm F_3}$ are added along the column respectively to generate two vectors of them when $w\geq h$. In the end, the vector of valid ${\rm F_3}$ is divided by the vector of $\frac{1}{4}$ GT shrink-mask to generate vector ${\rm F_s}$. The discriminator is composed of an LSTM~\cite{cho2014learning} based bidirectional RNN structure and a Fully Connect Network (FCN) based classifier. It abandons the zero regions of ${\rm F_s}$ and inputs the processed ${\rm F_s}$ into bidirectional RNN to extract two sequential features ${\rm hidden_1}$ and ${\rm hidden_2}$. The classifier treats the concatenation of ${\rm hidden_1}$ and ${\rm hidden_2}$ as input to estimate whether the region is shrink-mask, which encourages ZTD to extract the sequential feature and to combine it with the visual feature to suppress false-positive samples. Particularly, SVD is a training-only module, which brings no extra computational cost to the inference process and can be integrated into other detectors seamlessly.

\begin{figure*}
	\centering
	\subfigure[Original training smaples of MSRA-TD500]{
		\begin{minipage}[b]{0.2\linewidth}
			\includegraphics[width=1\linewidth]{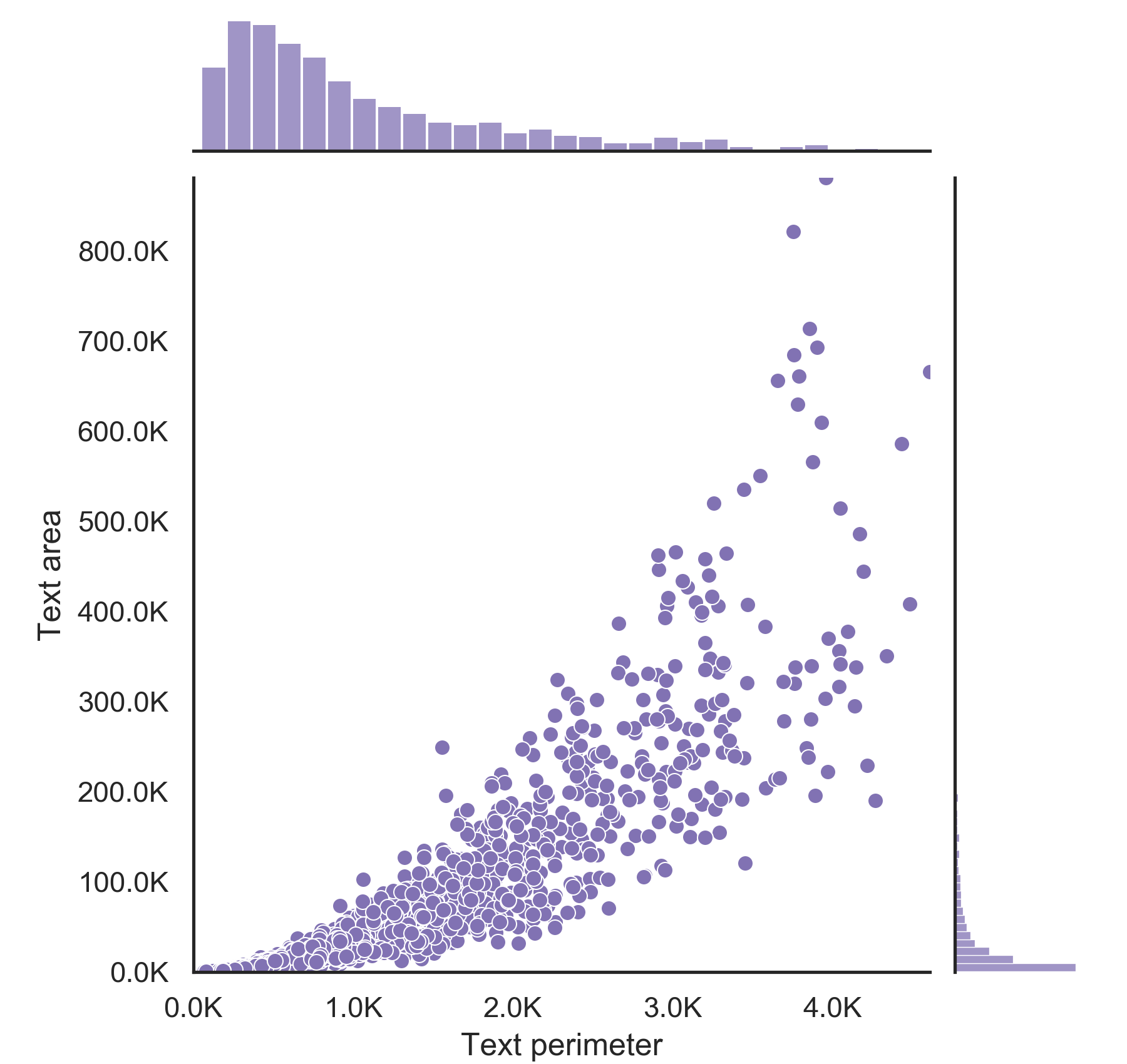}
	\end{minipage}}
	\subfigure[Original testing samples of MSRA-TD500]{
		\begin{minipage}[b]{0.2\linewidth}
			\includegraphics[width=1\linewidth]{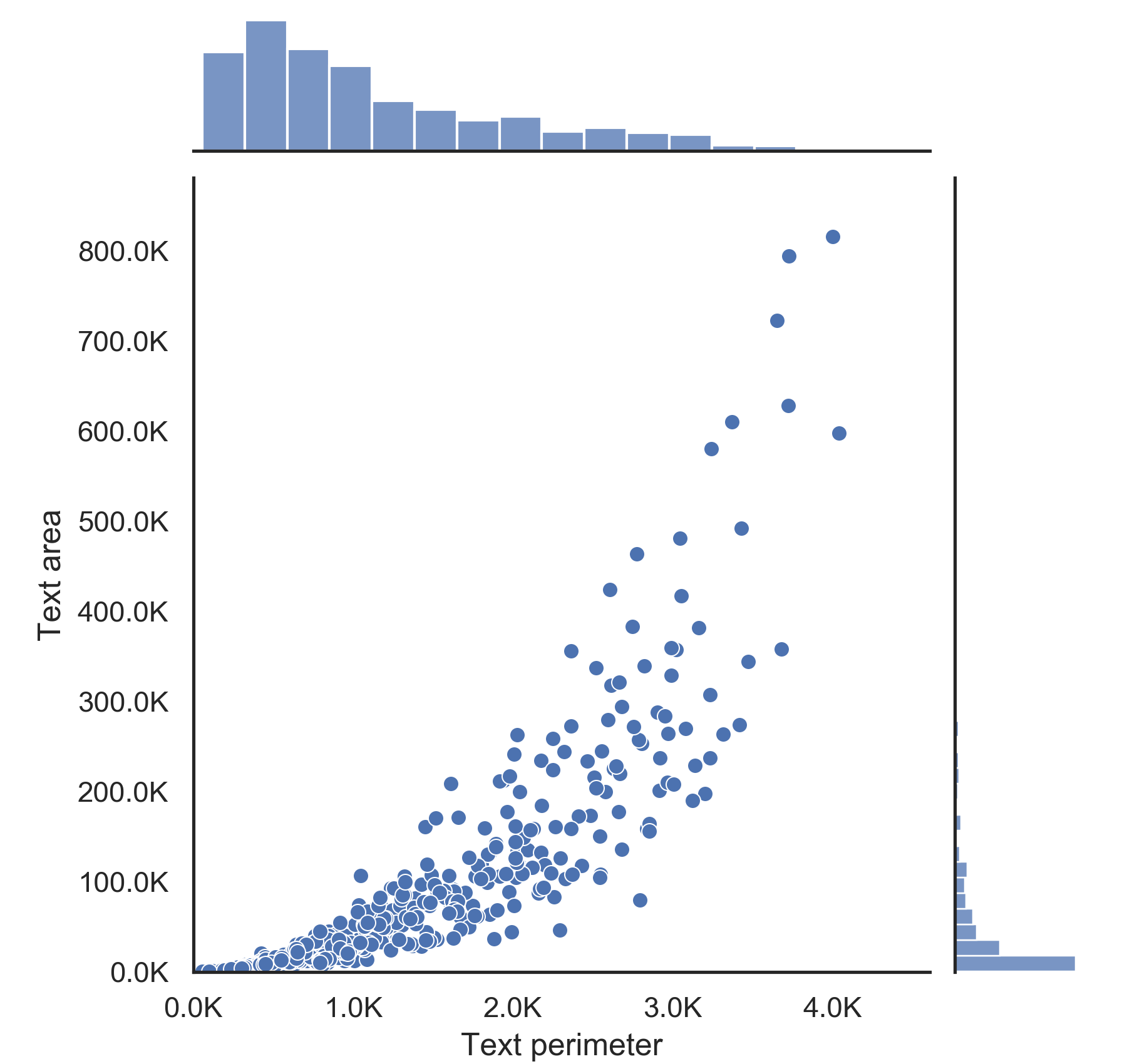}
	\end{minipage}}
	\subfigure[Resized training smaples of MSRA-TD500]{
		\begin{minipage}[b]{0.2\linewidth}
			\includegraphics[width=1\linewidth]{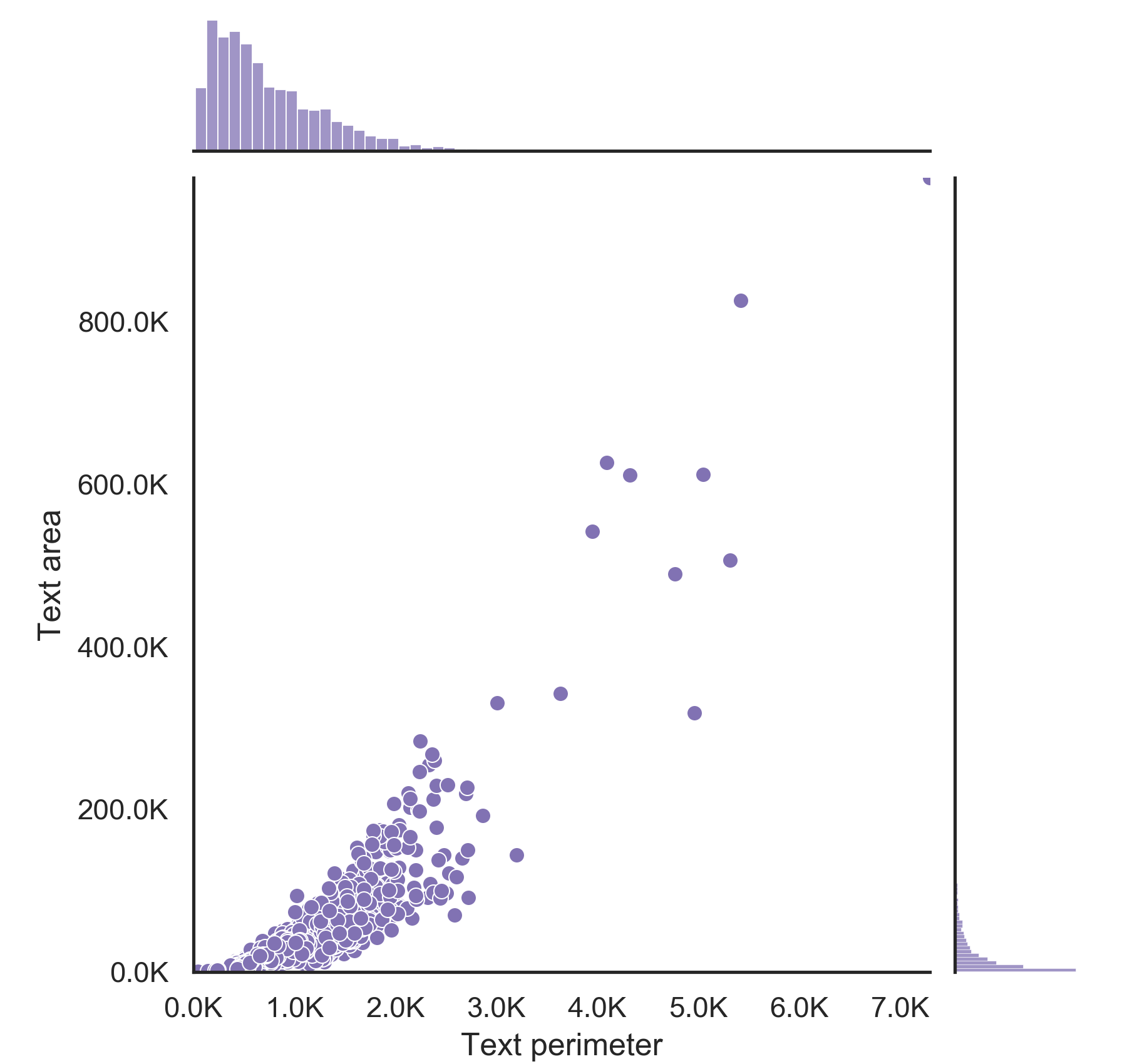}
	\end{minipage}}
	\subfigure[Resized testing samples of MSRA-TD500]{
		\begin{minipage}[b]{0.2\linewidth}
			\includegraphics[width=1\linewidth]{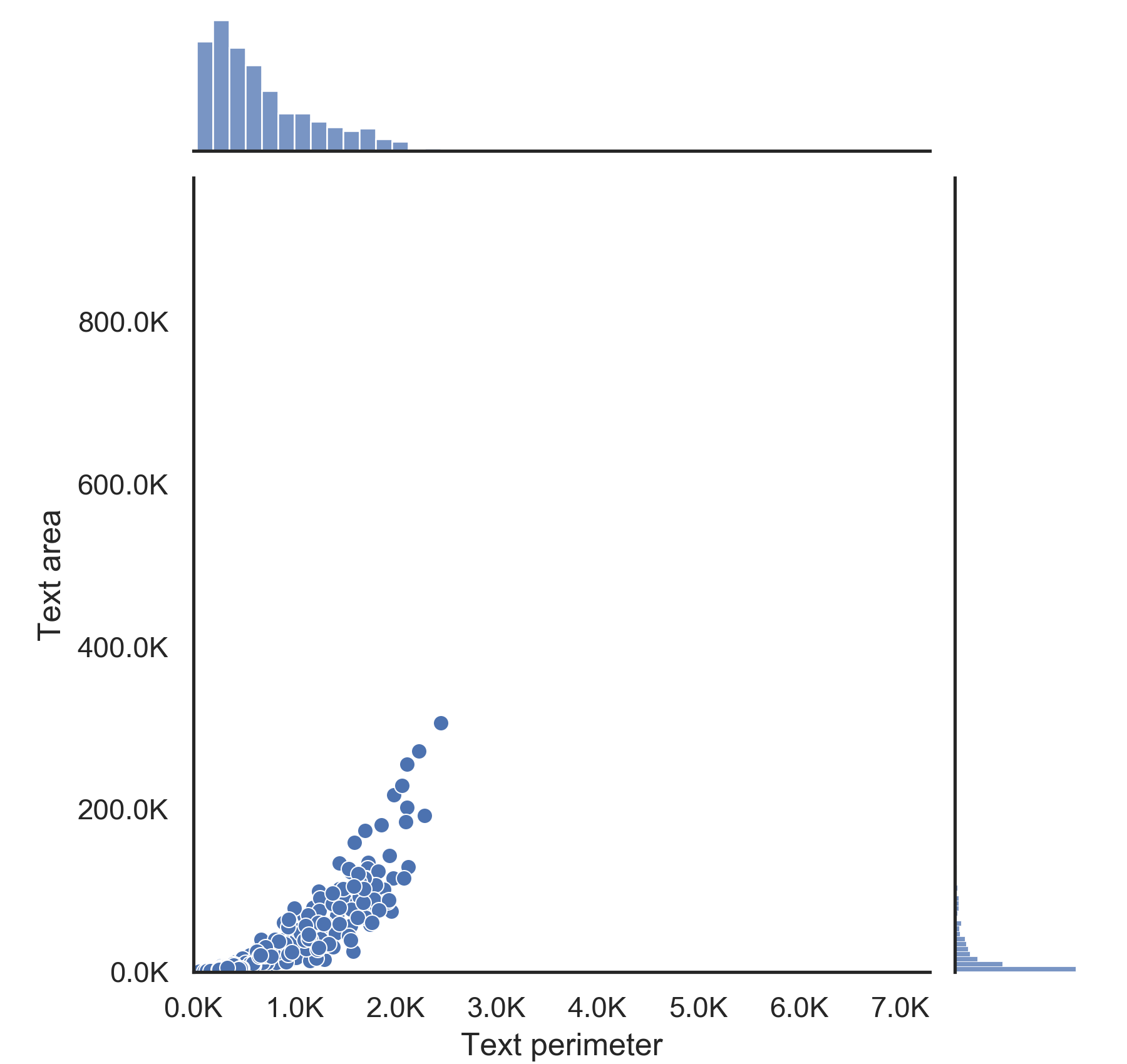}
	\end{minipage}}
	\vspace{-.10in}
	
	\subfigure[Original training smaples of Total-Text]{
		\begin{minipage}[b]{0.2\linewidth}
			\includegraphics[width=1\linewidth]{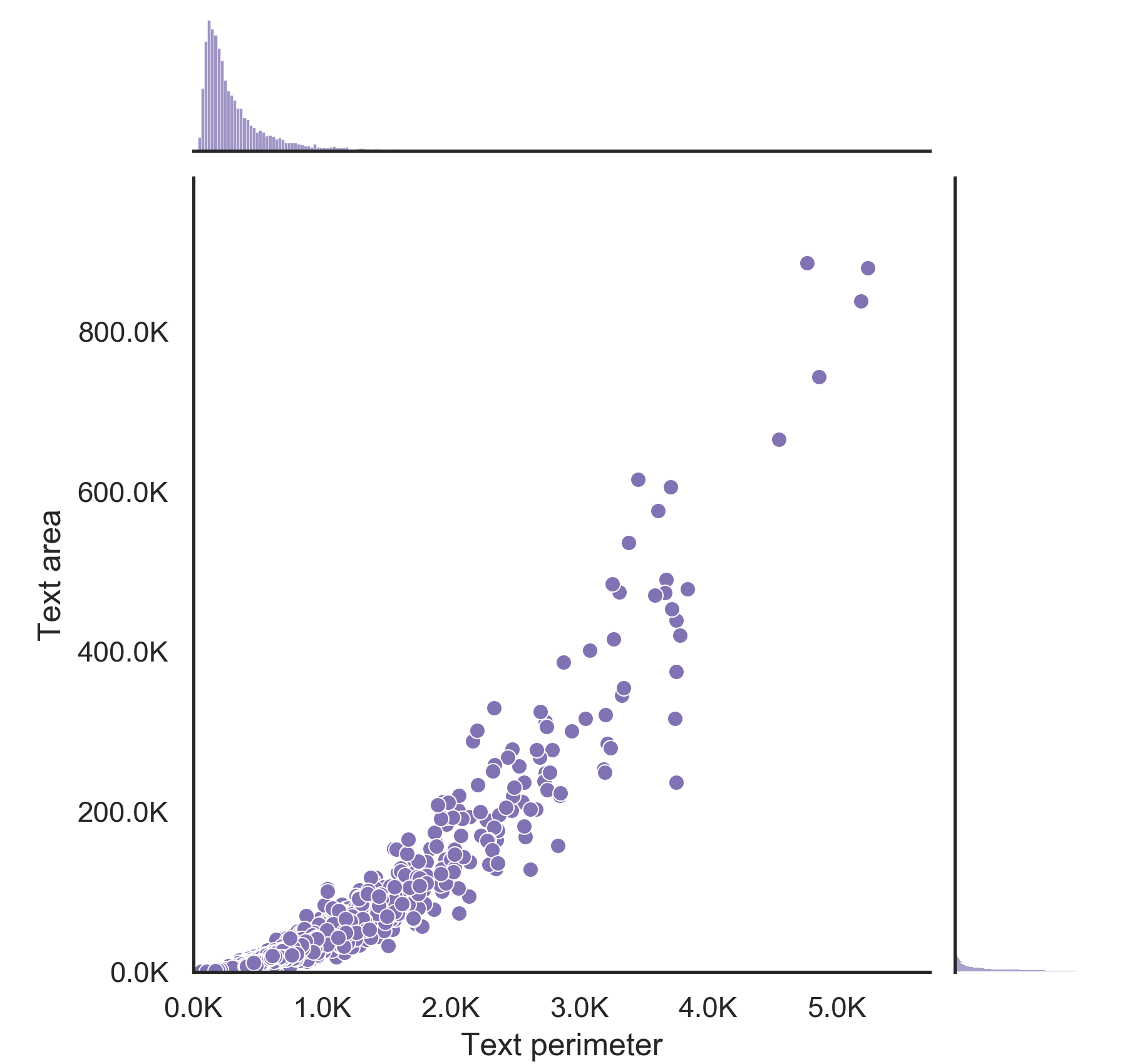}
	\end{minipage}}
	\subfigure[Original testing samples of Total-Text]{
		\begin{minipage}[b]{0.2\linewidth}
			\includegraphics[width=1\linewidth]{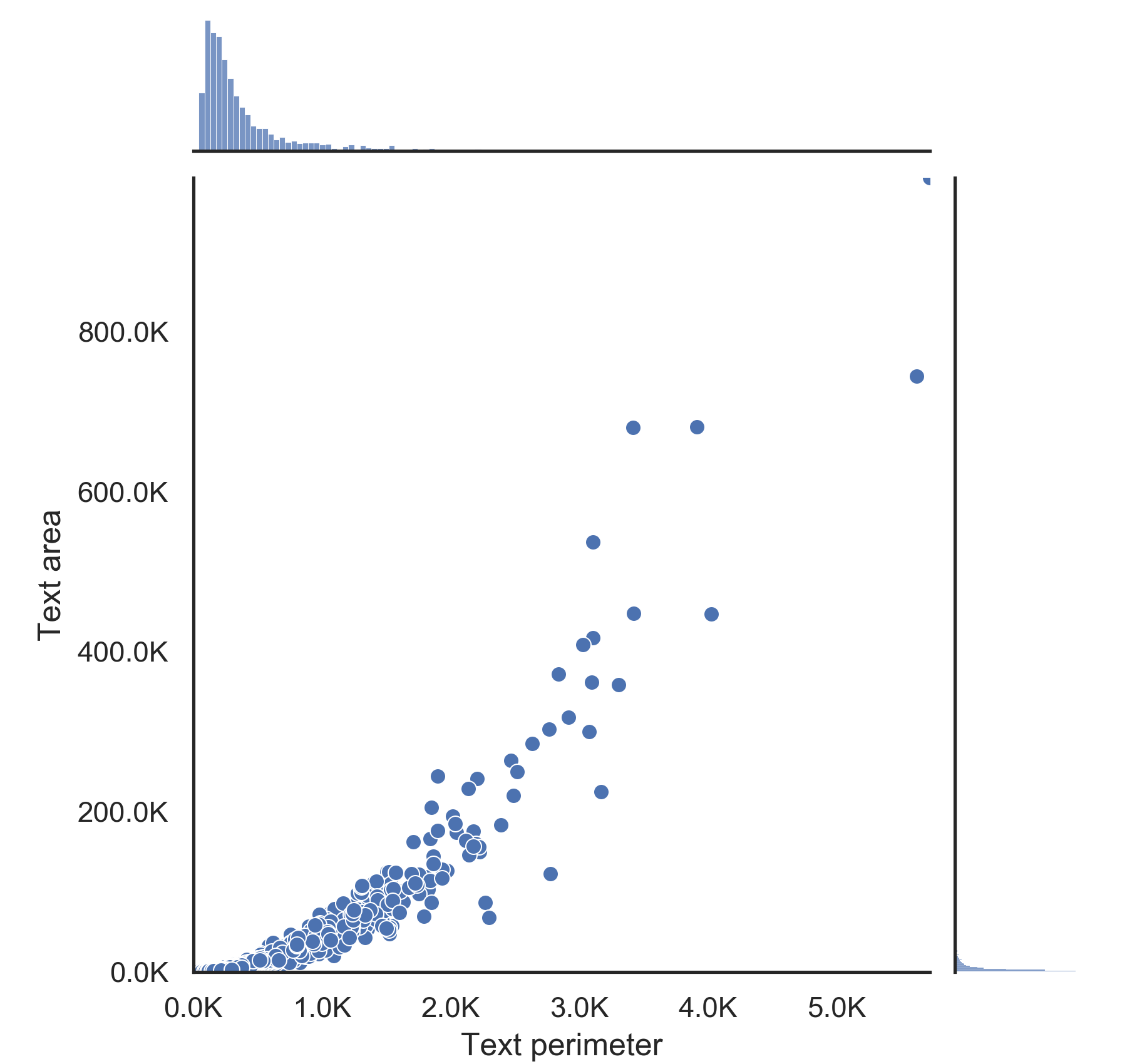}
	\end{minipage}}
	\subfigure[Resized training smaples of Total-Text]{
		\begin{minipage}[b]{0.2\linewidth}
			\includegraphics[width=1\linewidth]{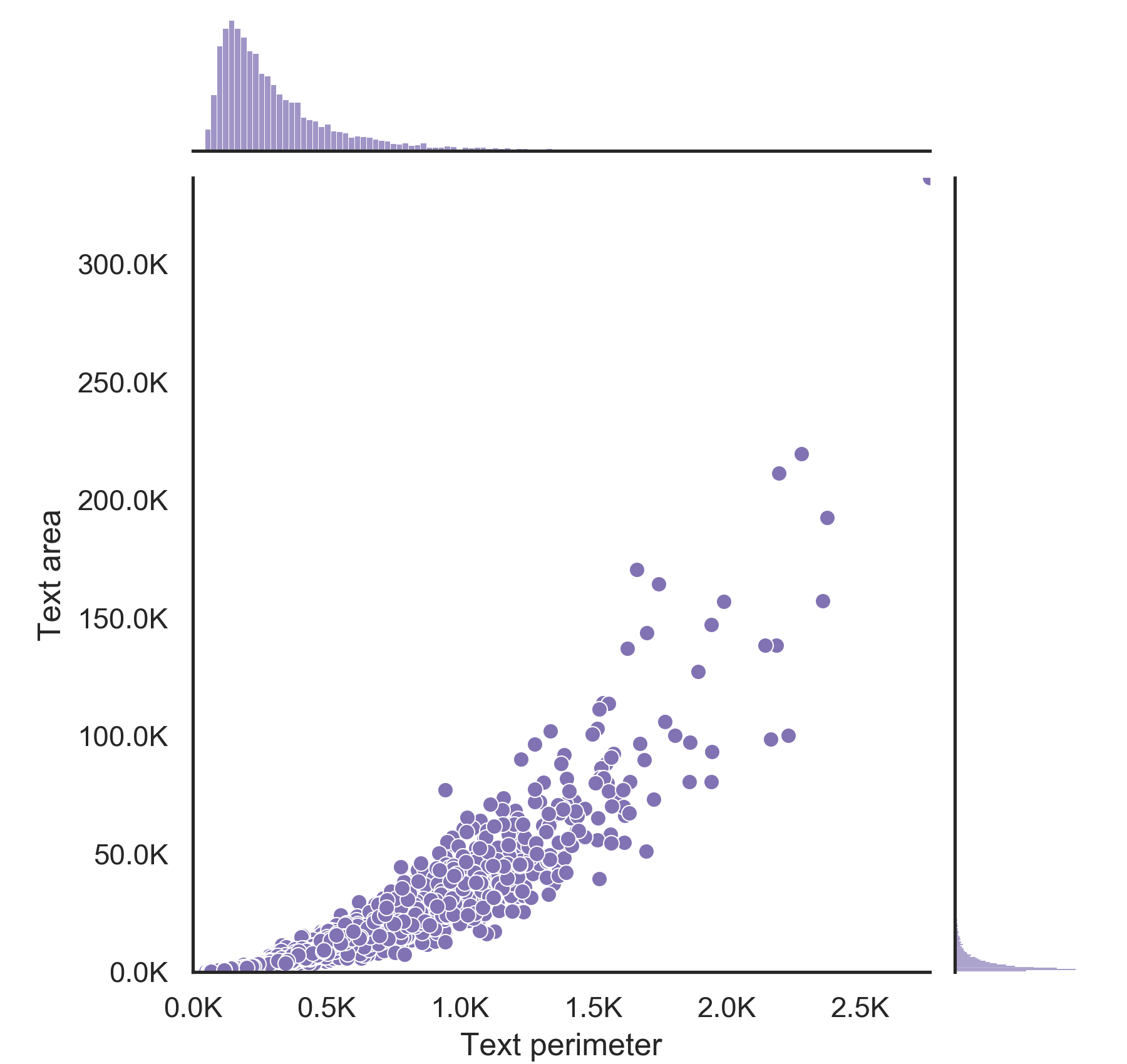}
	\end{minipage}}
	\subfigure[Resized testing samples of Total-Text]{
		\begin{minipage}[b]{0.2\linewidth}
			\includegraphics[width=1\linewidth]{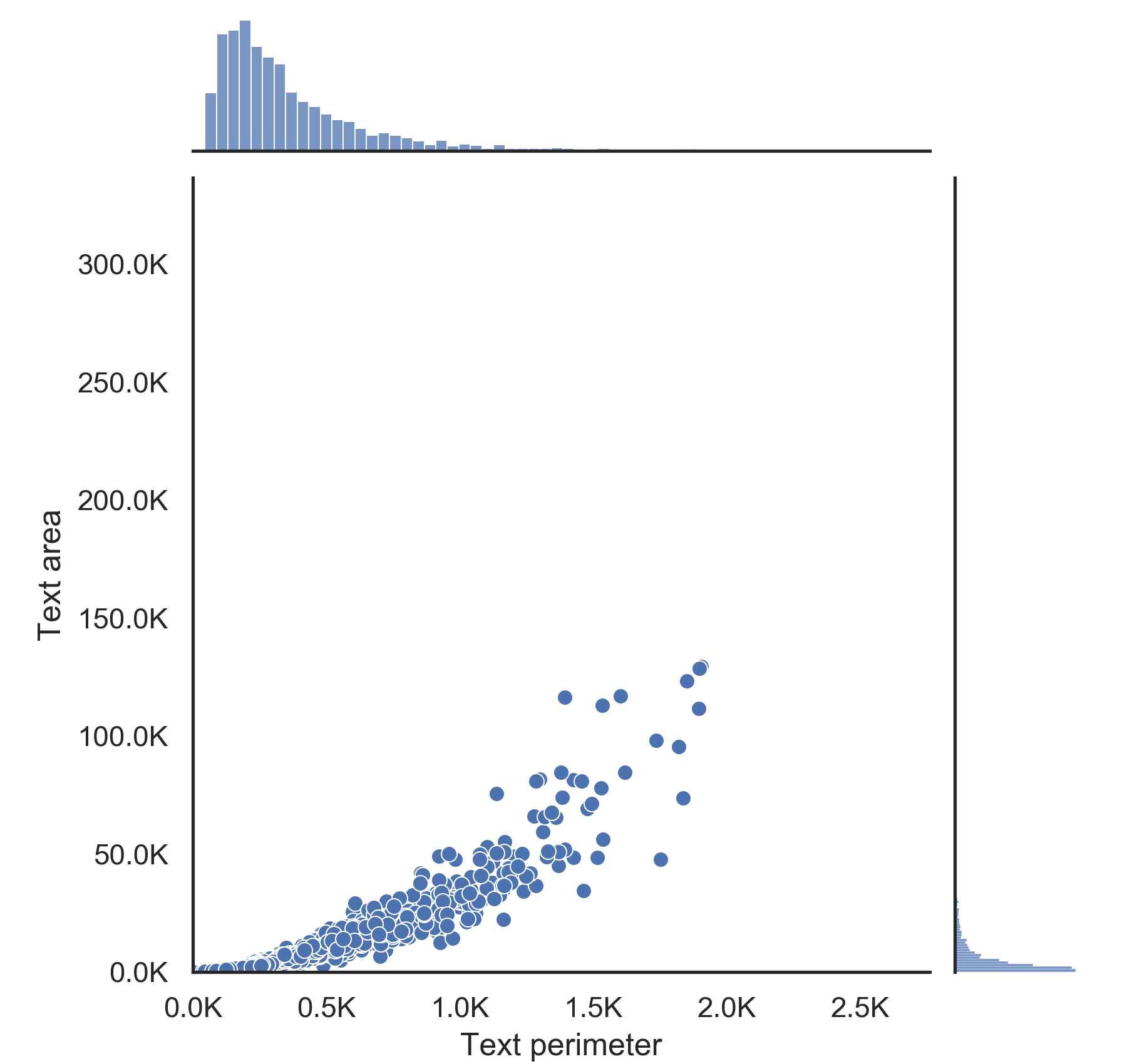}
	\end{minipage}}
	\vspace{-.10in}
	
	\subfigure[Original training smaples of CTW1500]{
		\begin{minipage}[b]{0.2\linewidth}
			\includegraphics[width=1\linewidth]{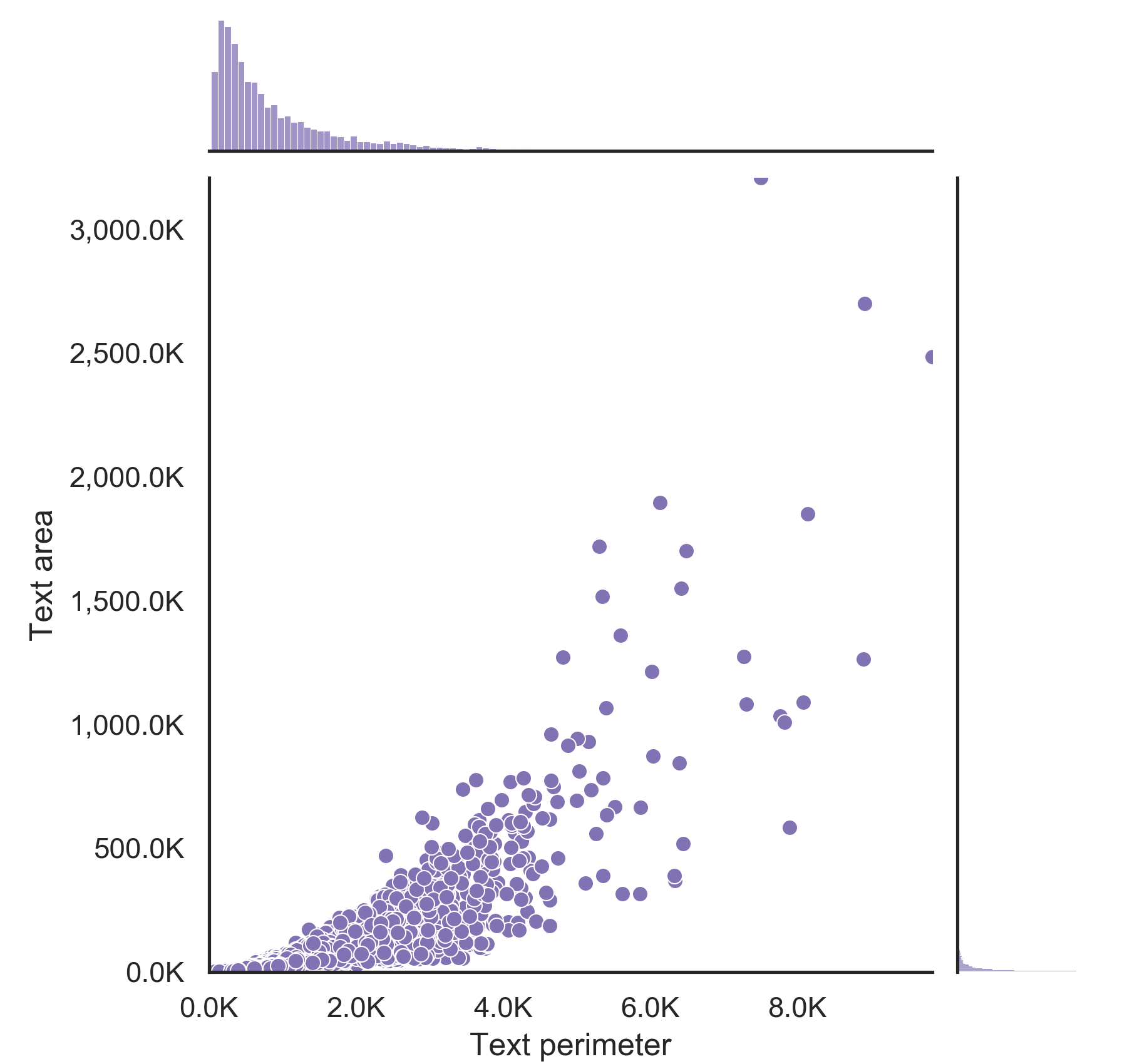}
	\end{minipage}}
	\subfigure[Original testing samples of CTW1500]{
		\begin{minipage}[b]{0.2\linewidth}
			\includegraphics[width=1\linewidth]{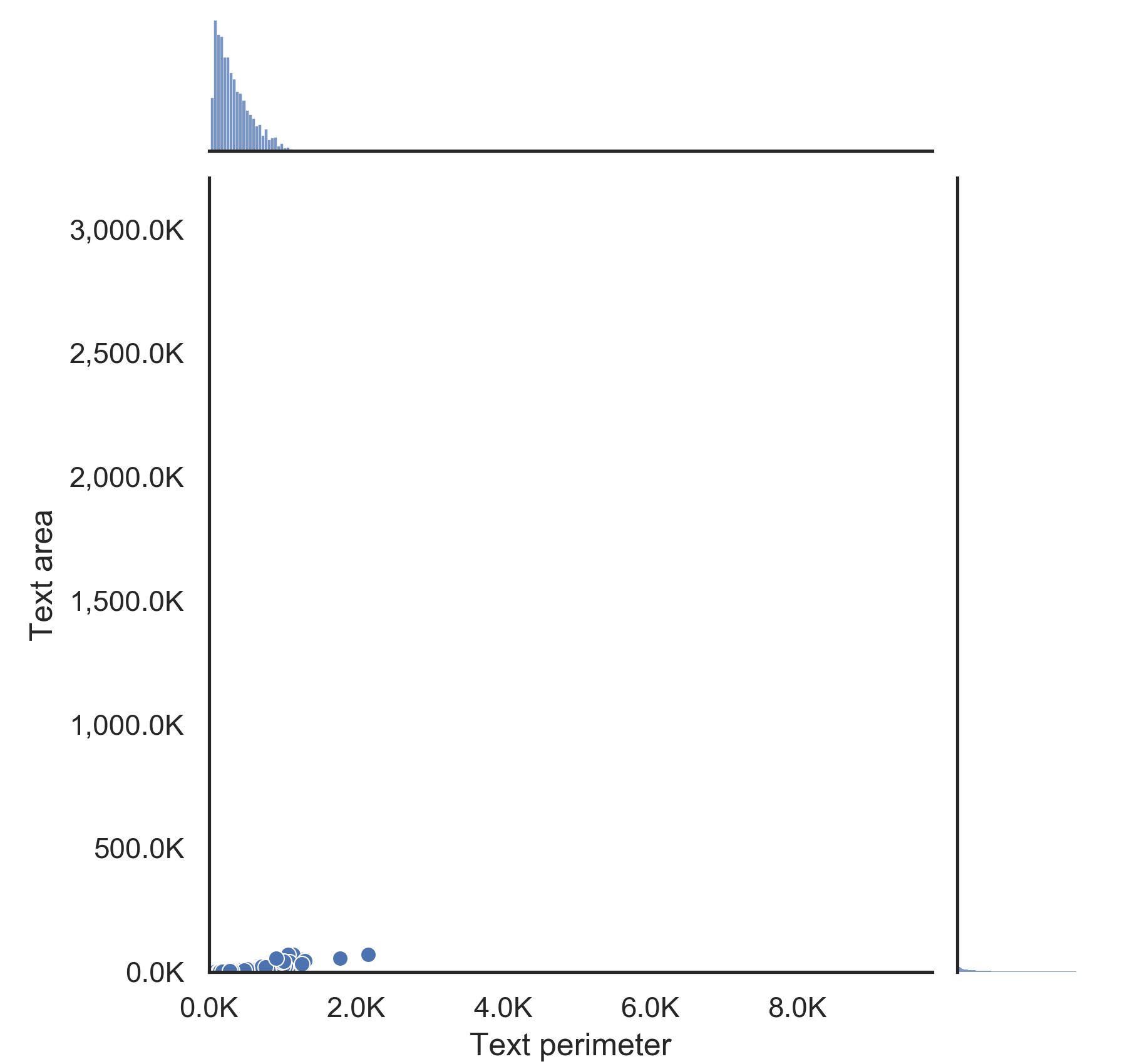}
	\end{minipage}}
	\subfigure[Resized training smaples of CTW1500]{
		\begin{minipage}[b]{0.2\linewidth}
			\includegraphics[width=1\linewidth]{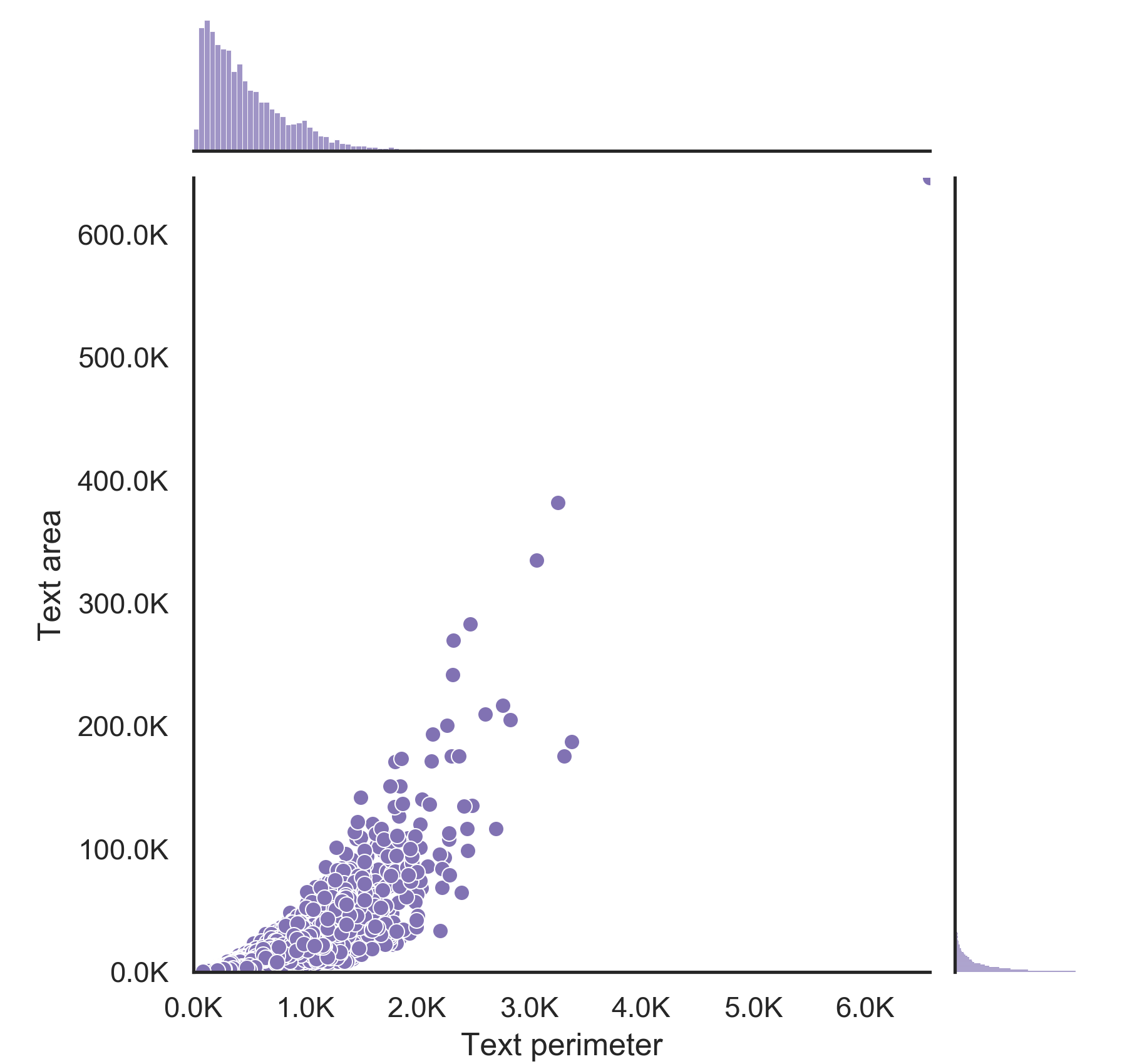}
	\end{minipage}}
	\subfigure[Resized testing samples of CTW1500]{
		\begin{minipage}[b]{0.2\linewidth}
			\includegraphics[width=1\linewidth]{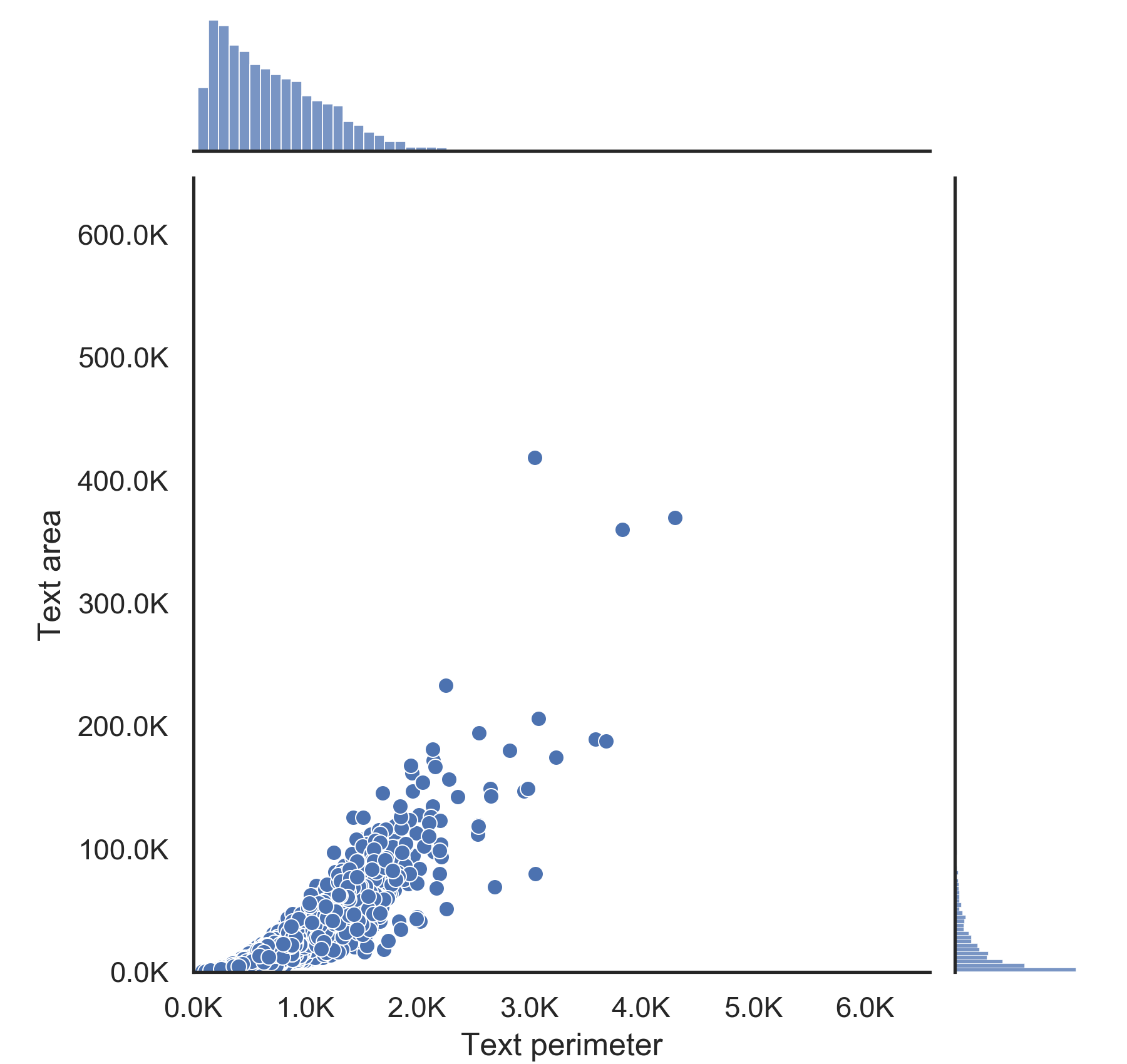}
	\end{minipage}}
	\vspace{-.10in}
	
	\subfigure[Original training smaples of ICDAR2015]{
		\begin{minipage}[b]{0.2\linewidth}
			\setlength{\abovecaptionskip}{0pt}
			\setlength{\belowcaptionskip}{0pt}
			\includegraphics[width=1\linewidth]{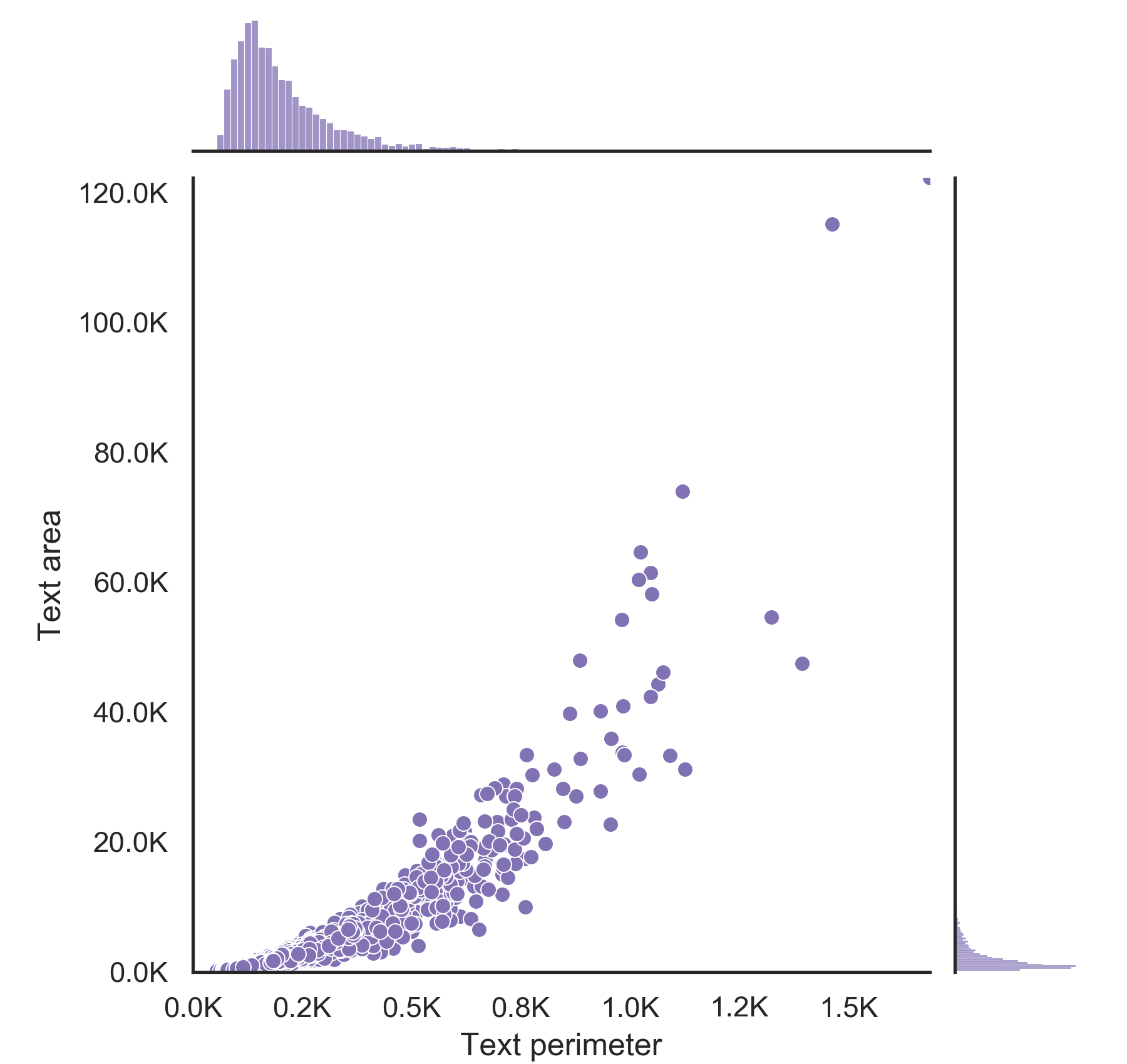}
	\end{minipage}}
	\subfigure[Original testing samples of ICDAR2015]{
		\begin{minipage}[b]{0.2\linewidth}
			\includegraphics[width=1\linewidth]{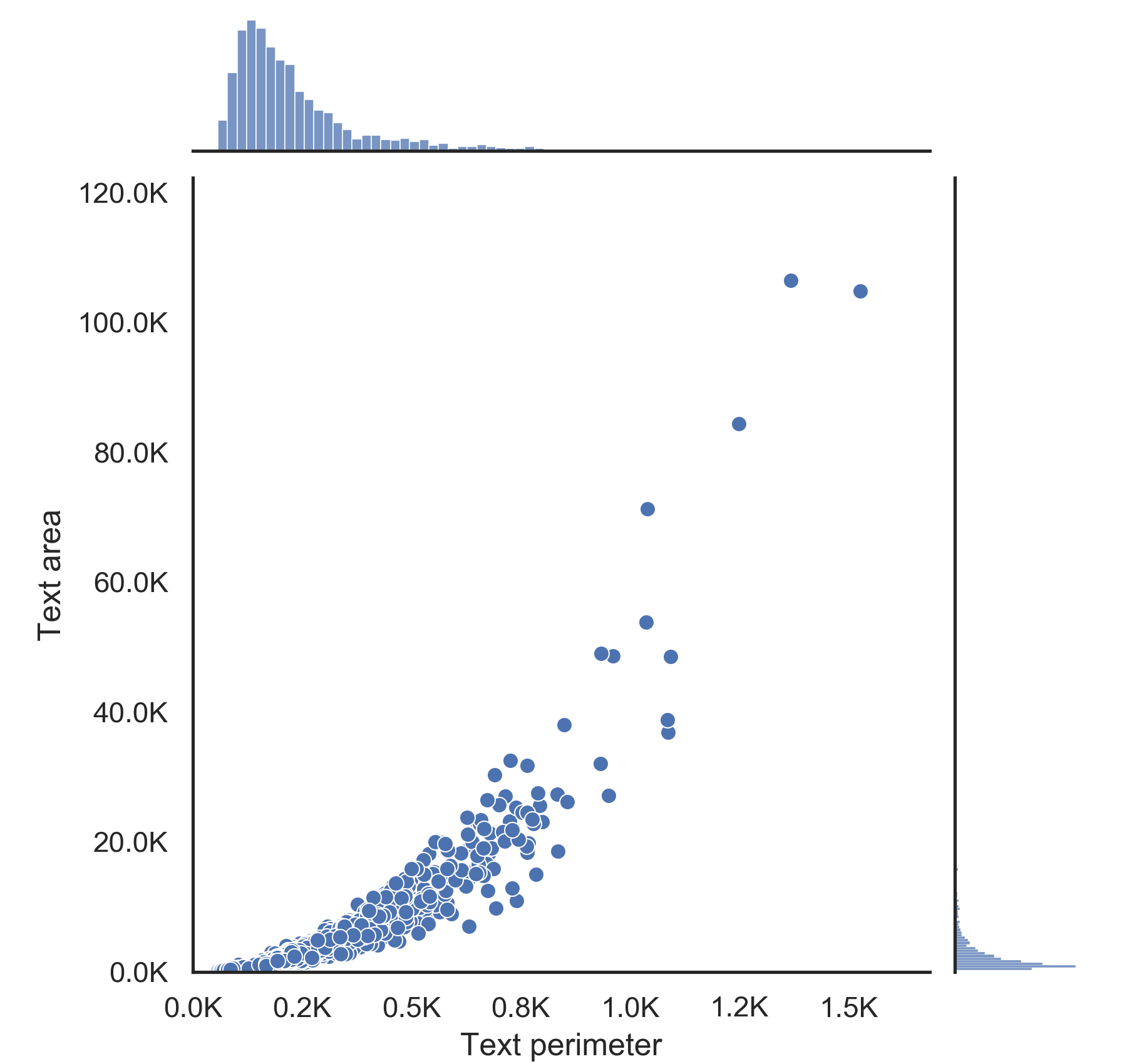}
	\end{minipage}}
	\subfigure[Resized training smaples of ICDAR2015]{
		\begin{minipage}[b]{0.2\linewidth}
			\includegraphics[width=1\linewidth]{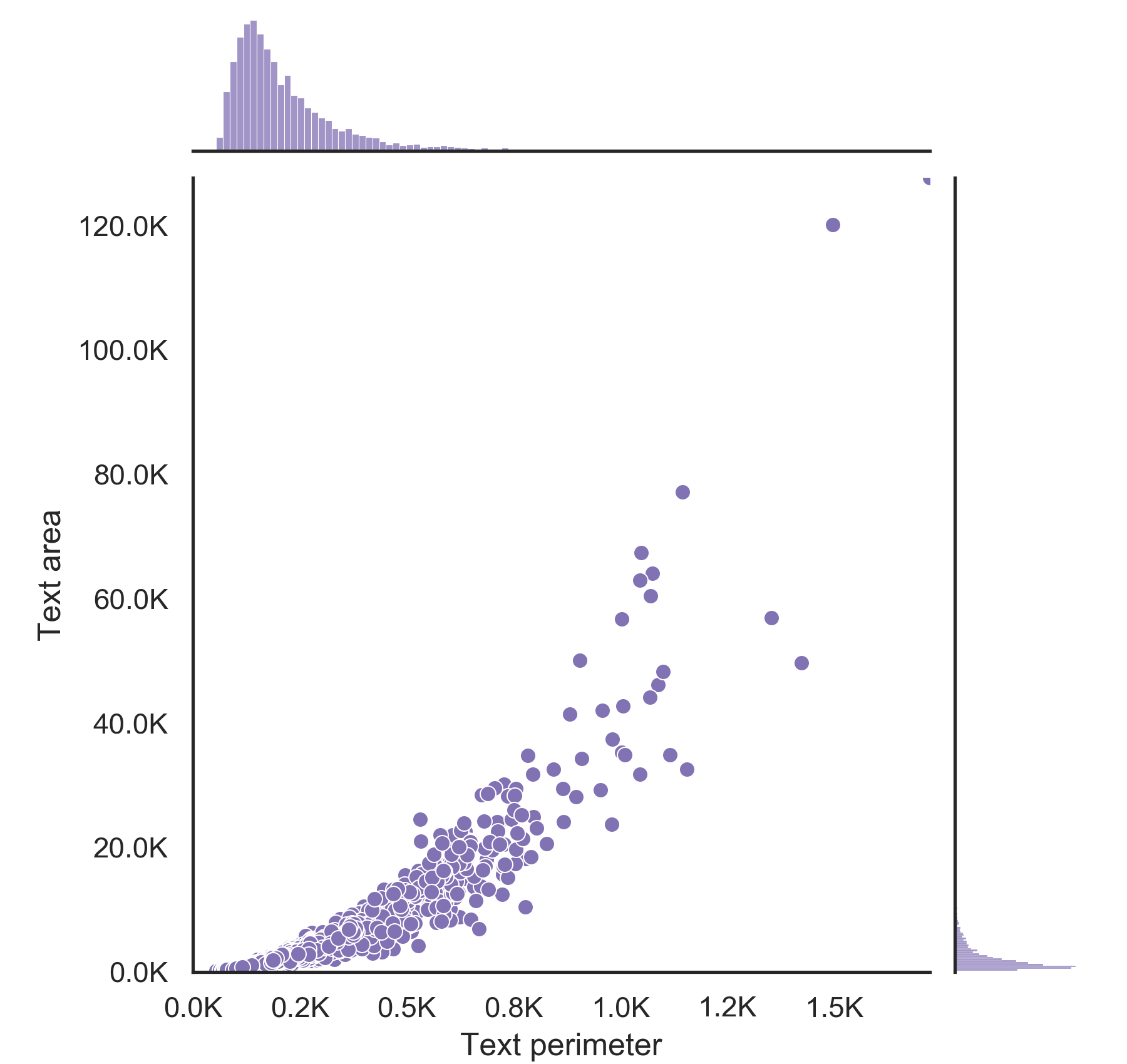}
	\end{minipage}}
	\subfigure[Resized testing samples of ICDAR2015]{
		\begin{minipage}[b]{0.2\linewidth}
			\includegraphics[width=1\linewidth]{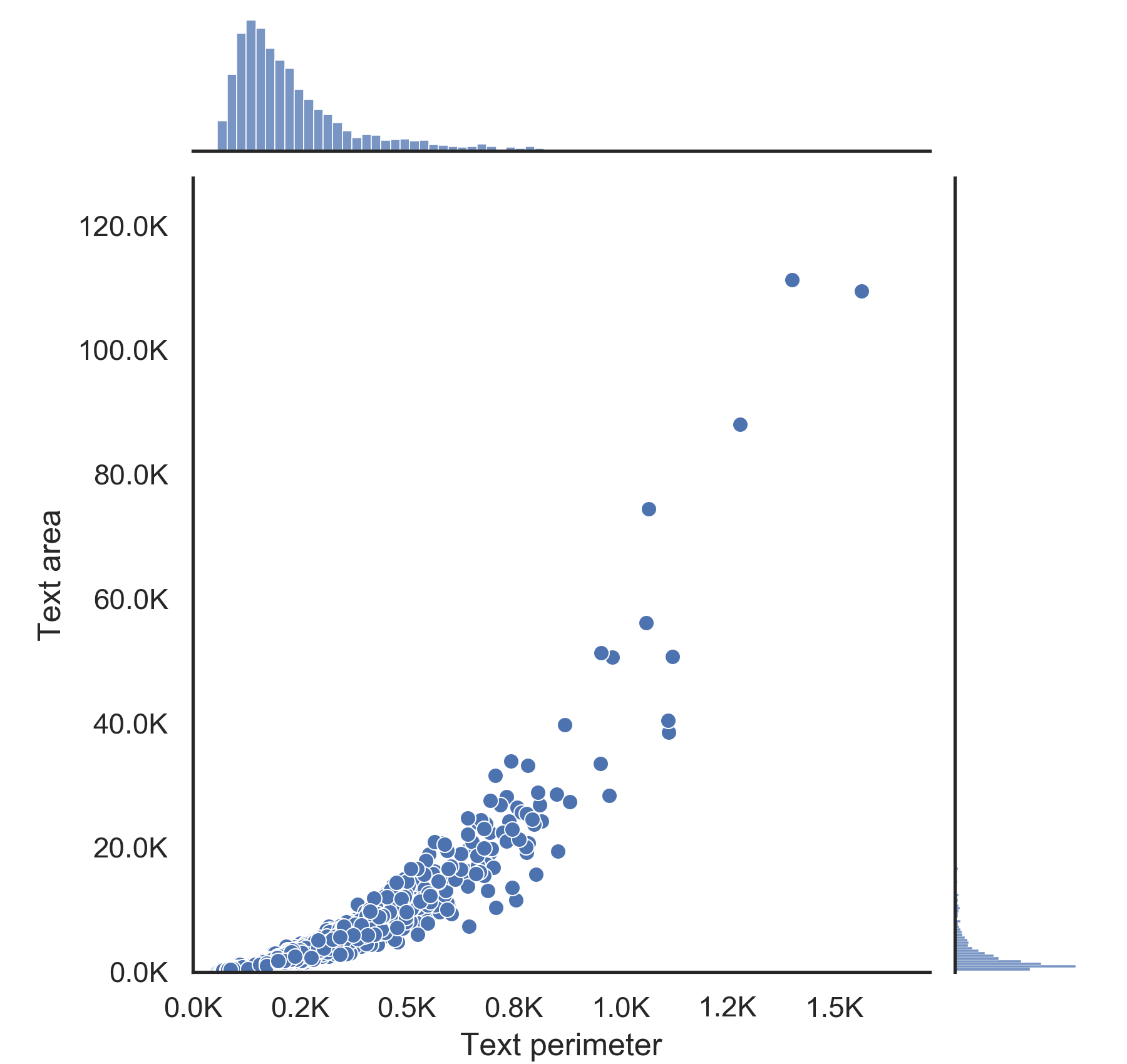}
	\end{minipage}}
	
	\caption{Visualization of geometry characteristics of text instances in MSRA-TD500~(the first row), Total-Text~(the second row), CTW1500~(the third row), and ICDAR2015~(the fourth row) datasets.}
	\label{dataset}
\end{figure*}

\subsection{Loss Function}
\label{lossfunction}
As we can see from Fig.~\ref{V2}, the proposed ZTD is composed of shrink-mask prediction header, ZIM, ZOM, and SVD. Therefore, the overall objective function consists of ${\cal L}_{sm}$,  ${\cal L}_{zi}$, ${\cal L}_{zo}$, and ${\cal L}_{svd}$, which can be formulated as:
\begin{eqnarray}
{\cal L}=\alpha {\cal L}_{sm}+\beta {\cal L}_{zi} +\gamma {\cal L}_{zo} +\eta {\cal L}_{svd},
\end{eqnarray}          
where the parameters $\alpha$, $\beta$, $\gamma$, and $\eta$ balance the importance of different loss functions. They are set to 1, 0.25, 0.25, and 0.25 in the following experiments, respectively.

\textbf{Optimization of shrink-mask prediction header.} Dice loss~\cite{milletari2016v} is proposed to evaluate the similarity of different binary masks. Particularly, it performs better than other loss functions when positive and negative samples are imbalanced, which is suitable for the shrink-mask prediction task. Therefore, we adopt the dice loss to evaluate the loss ${\cal L}_{sm}$ of this header, which is defined as:
\begin{eqnarray}
{\cal L}_{sm} = 1-\frac{2\times| {\rm SM_{p}}\cap {\rm SM_{g}}|+\varepsilon}{| {\rm SM_{p}} |+| {\rm SM_{g}}|+\varepsilon},
\end{eqnarray}  
where ${\rm SM_{p}}$ and ${\rm SM_{g}}$ indicate the predicted shrink-mask and the corresponding ground-truth. Considering that there may be no positive samples in ground-truth, we set $\varepsilon$ as 1 to avoid the denominator equal to 0.

\textbf{Optimization of Zoom In Module.} As we mentioned before, the shrink-mask is generated by shrinking the text contour inward by a specific distance, which means both the margins and shrink-masks are parts of texts. It makes existing methods hard to discriminate them, which may lead to ambiguous shrink-mask edges and further influence model performance. To recognize the edges accurately, we focus on the margins through ZIM. The loss function ${\cal L}_{zi}$ of ZIM can be expressed as:
\begin{eqnarray}
{\cal L}_{zi} = 1-\frac{2\times| ({\rm ZI_{p}})\cap {\rm ZI_{g}}|+\varepsilon}{| ({\rm ZI_{p}}) |+| {\rm ZI_{g}}|+\varepsilon},
\end{eqnarray}  
where ${\rm ZI_{p}}$ denotes the predicted binary mask of the margin and ${\rm ZI_{g}}$ is the corresponding label.

\textbf{Optimization of Zoom Out Module.} The label of this module is shrink-mask of $\frac{1}{16}$ stride, which can be generated by the Label Generation Process in Fig.~\ref{V4}. The same as shrink-mask prediction header, dice loss is used for the evaluation of the loss ${\cal L}_{zo}$ between the predicted binary mask and ground-truth:
\begin{eqnarray}
{\cal L}_{zo} = 1-\frac{2\times| {\rm ZO_{p}}\cap {\rm ZO_{g}}|+\varepsilon}{| {\rm ZO_{p}} |+| {\rm ZO_{g}}|+\varepsilon},
\end{eqnarray}  
where ${\rm ZO_{p}}$ and ${\rm ZO_{g}}$ are the predicted coarse shrink-mask and the corresponding ground-truth.

\textbf{Optimization of Sequential-Visual Discriminator.} Considering false-positive samples enjoy similar visual features with texts, SVD is presented to encourage our model to suppress them by the combination of sequential and visual features. For this classification task, we adopt BCE loss to measure the loss ${\cal L}_{svd}$ of this module:
\begin{eqnarray}
{\cal L}_{svd} = -({\rm S_p}\times{\rm log}({\rm S_g})+(1-{\rm S_p})\times{\rm log}(1-{\rm S_g}),
\end{eqnarray}  
where ${\rm S_p}$ is the probability whether the region is shrink-mask and ${\rm S_g}$ denotes the ground-truth.

\section{Experiments}
\label{sec4}
\subsection{Datasets}
To verify the effectiveness and robustness of our method to the texts with different shapes, scales, and aspect ratios, we evaluate ZTD on the four representative public datasets:

\textbf{MSRA-TD500}~\cite{yao2012detecting} is a dataset consisting of line-level Chinese and English text instances. There are 300 training images and 200 testing images, respectively. The same as previous works, we introduce 400 extra images from HUST-TR400~\cite{yao2014unified} as training data.

\textbf{Total-Text}~\cite{ch2017total} contains horizontal, multi-oriented, curved and other irregular-shaped texts. Except for English texts, there are still some Chinese and Japanese samples, which brings difficulty for detection. This dataset contains 1255 training images and 300 testing images, respectively.

\textbf{CTW1500}~\cite{yuliang2017detecting} has 1000 training images and 500 testing images. Different from Total-Text, this dataset mainly consists of line-level arbitrary-shaped text instances.

\textbf{ICDAR2015}~\cite{karatzas2015icdar} is proposed in ICDAR 2015 Robust Reading Competition. Compared with the above three public benchmarks, ICDAR2015 has a more complicated background, which makes it hard to distinguish text and interference region. The same as CTW1500, ICDAR2015 utilizes 1000 images to train the model and 500 images to evaluate the detection performance.

\begin{table*}[]
	\renewcommand{\arraystretch}{1.1}
	\setlength{\tabcolsep}{3mm}
	\centering
	\caption{Detection results of ZTD with different settings on MSRA-TD500. ``S: 736'' means that the short side of each testing image is resized to be 736 pixels. ``baseline'' means the framework equipped with shrink-mask prediction header only. ``Ext.'' indicates that ZTDl is pre-trained on SynthText~\cite{gupta2016synthetic}.}
	\begin{tabular}{clcccccccc}
		\toprule
		&           \multicolumn{9}{c}{Image scale for testing~ (S~:~736)}     \\ \midrule
		\# & Methods    & ZIM & ZOM & SVD & Ext. & P & R & F & FPS \\ \midrule
		1  & baseline  &         &               &            & & 86.4 & 79.9 & 83.0 & 64.1   \\ 
		2  & baseline+ & \checkmark &            &            & & 87.5 & 80.5 & 83.9 & 64.1    \\ 
		3  & baseline+ & \checkmark & \checkmark &            & & 90.7 & 80.3 & 85.2 & 59.2    \\ 
		4  & baseline+ & \checkmark & \checkmark & \checkmark & & 92.2 & 80.9 & 86.2 & 59.2     \\ 
		5  & baseline+ & \checkmark & \checkmark & \checkmark & \checkmark & 91.6 & 82.4 & 86.8 & 59.2    \\ \bottomrule
	\end{tabular}
	\label{t1}
\end{table*}

\begin{figure*}
	\begin{center}		
		\includegraphics[width=0.9\textwidth]{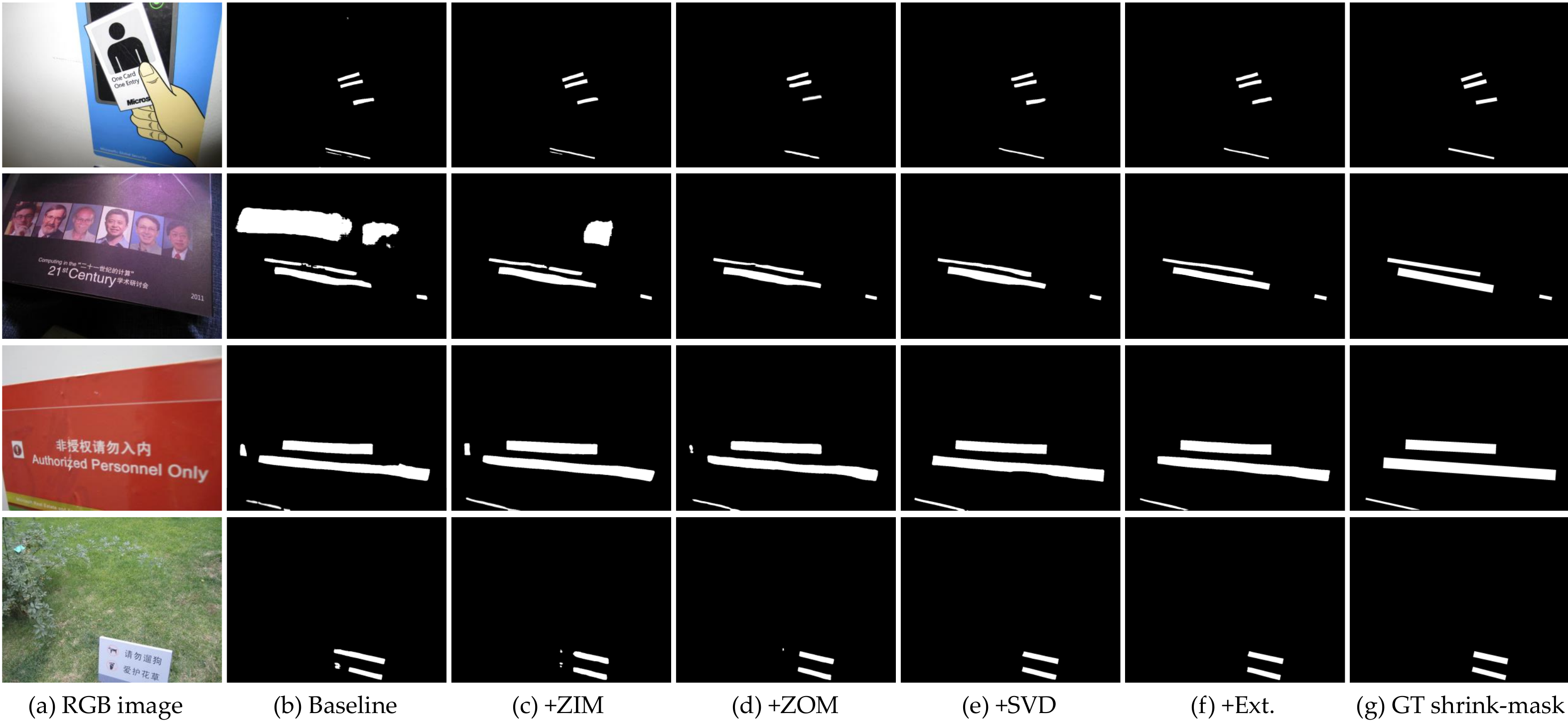}
	\end{center}
	\vspace{-5mm}
	\caption{Visualization of the predicted shrink-masks of ZTD with different settings.}
	\label{R1}
\end{figure*}

The geometry characteristics of text instances of different datasets are shown in Fig.~\ref{dataset}. For the original datasets (as illustrated in the first and second columns in Fig.~\ref{dataset}), text scales of CTW1500 are almost 25 times bigger than the text of ICDAR2015. Moreover, there are huge characteristic differences between the training and testing text instances of CTW1500, which brings difficulty for text detection. For the text instances of MSRA-TD500 and Total-Text, they enjoy similar geometry characteristics. To ensure a fair comparison environment, we resize the short sides of original images to specific sizes to generate resized text instances. It can be found from the third and fourth columns in Fig.~\ref{dataset} that the resized training texts and testing texts enjoy similar geometry characteristic distributions.

\subsection{Implementation Details}
The overall architecture of our method is shown in Fig.~\ref{V2}, where the feature maps ($f_1$, $f_2$, $f_3$, and $f_4$) behind input are generated by different stages (stage1, stage2, stage3, and stage4 respectively) of ResNet-18~\cite{he2016deep}. 

In the data pre-processing stage, the training samples are increased by the following augmentation strategies: (1) random scaling (including image size and aspect); (2) random horizontal flipping; (3) random rotating in the range of (-10°, 10°); (4) random cropping and padding.

In the initializing stage, the backbone of ZTD is pre-trained on ImageNet~\cite{deng2009imagenet} and the rest of the layers are initialized by the strategy proposed in~\cite{he2015delving}. In the training process, the Adam~\cite{kingma2014adam} is deployed to optimize the model. For learning rate, it is initialized as 0.001 and adjusted through 'polylr' strategy. In the following experiments, our model is pre-trained on the SynthText dataset for 1 epoch and finetuned on the corresponding real-world datasets for 1200 epochs. The training batch size is set to 16. Moreover, the text instances labeled as DO NOT CARE are ignored during both training and testing stages. In the inference process, the red flows in Fig.~\ref{V2}, Fig.~\ref{V4}, Fig.~\ref{V5}, and Fig.~\ref{V6} are abandoned, which is helpful to facilitate detection speed. All the experiments in this paper are performed on a workstation with 1080Ti GPU.

\subsection{Ablation Study}
To verify the effectiveness of the proposed ZIM, ZOM, and SVD, we conduct an ablation study in this section. Furthermore, we explore the impacts of each sub-loss of ${\cal L}$ and the importances of different RNN units of SVD, respectively. The details of experimental results are described in the following paragraphs.

\begin{figure}
	\centering
	\subfigure[Details of training loss]{
		\begin{minipage}[b]{0.45\linewidth}
			\includegraphics[width=1\linewidth]{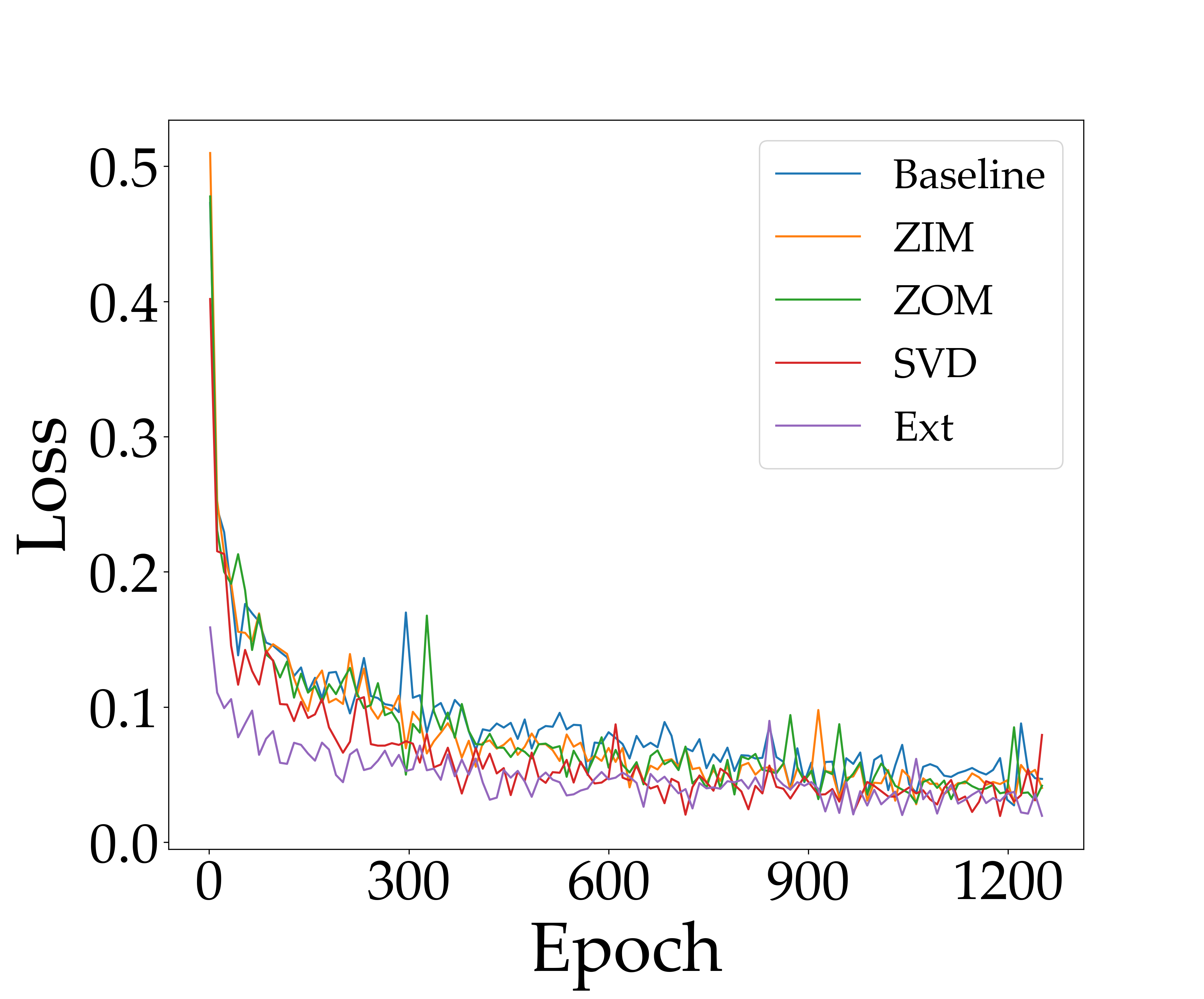}
	\end{minipage}}
	\subfigure[Details of training IoU]{
		\begin{minipage}[b]{0.45\linewidth}
			\includegraphics[width=1\linewidth]{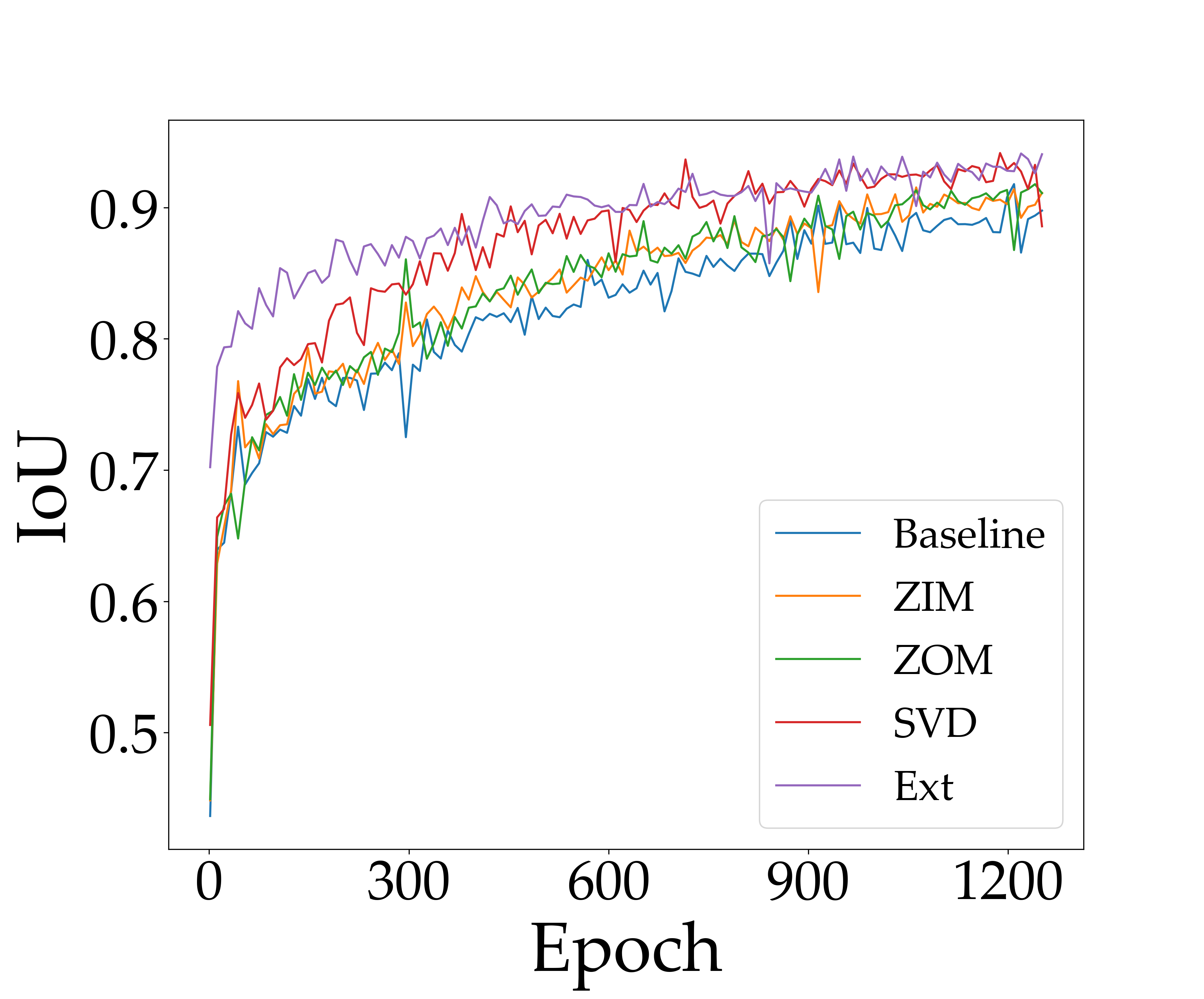}
	\end{minipage}}
	
	\caption{Convergence analysis of ZTD with different settings in Table~\ref{t1} on MSRA dataset. `IoU' means the Intersection of Union between predicted shrink-mask and the corresponding ground-truth.}
	\label{R2}
\end{figure}

\begin{figure}
	\centering
	\subfigure[Impact of different RNN units for detection accuracy]{
		\begin{minipage}{0.4\linewidth}
			\renewcommand{\arraystretch}{1.1}
			\setlength{\tabcolsep}{2mm}
			\footnotesize
			\begin{tabular}{cccc}
				\toprule
				Unit & P    & R    & F    \\ \midrule
				RNN  & 89.7 & 81.4 & 85.3 \\ \hline
				GRU  & 90.5 & 81.5 & 85.8 \\ \hline
				LSTM & 92.2 & 80.9 & 86.2 \\ \bottomrule
			\end{tabular}
	\end{minipage}}
	\hspace{.20in}
	\subfigure[Impact of the number of RNN layers for detection accuracy]{
		\begin{minipage}{0.45\linewidth}
			\includegraphics[width=1\linewidth]{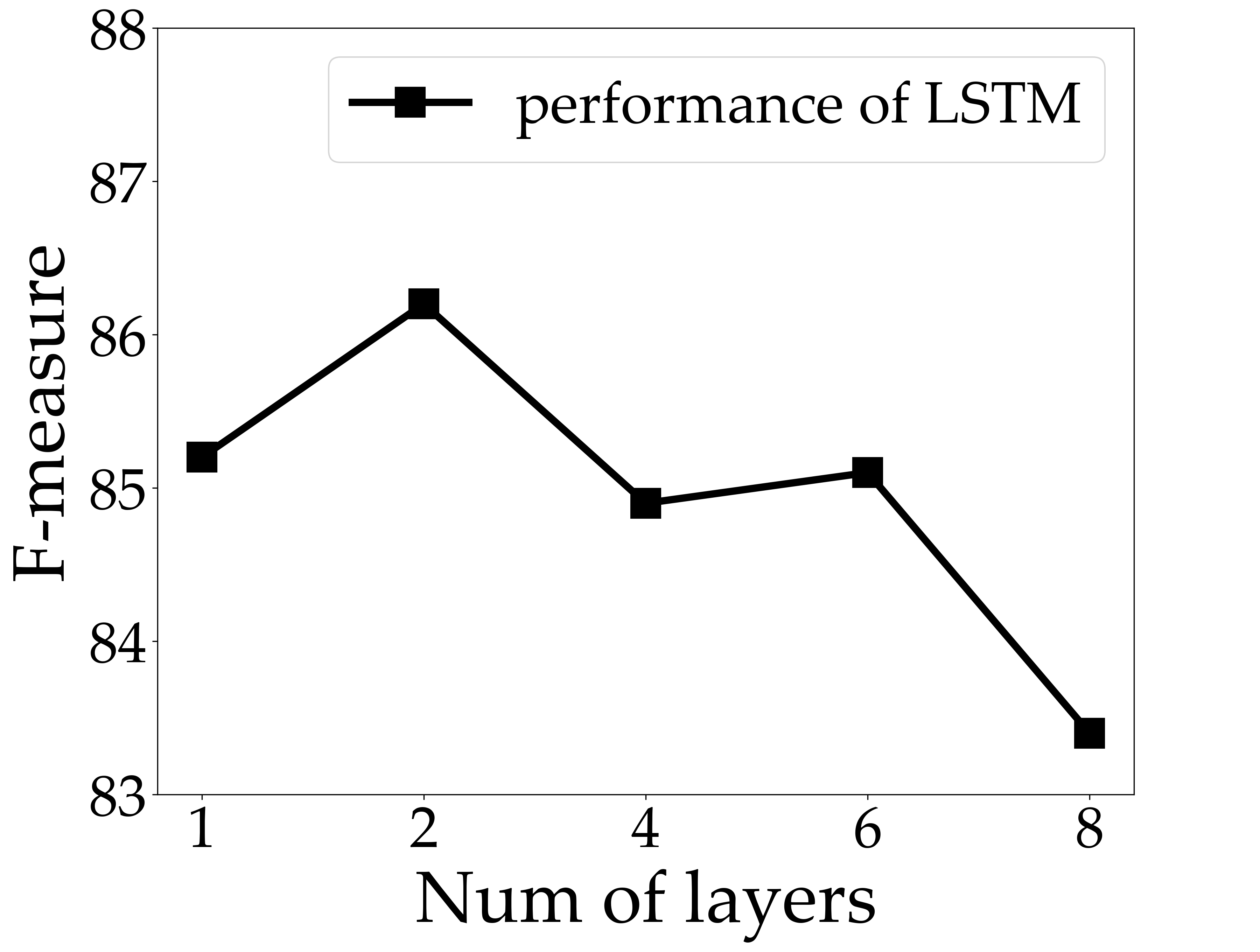}
	\end{minipage}}
	\caption{Detection results of SVD with different settings on MSRA-TD500. 'Unit` indicates the unit of RNN structure in SVD.}
	\label{t2}
\end{figure}

\begin{figure}
	\centering
	\subfigure[Impact of $\beta$ for F-measure]{
		\begin{minipage}[b]{0.45\linewidth}
			\includegraphics[width=1\linewidth]{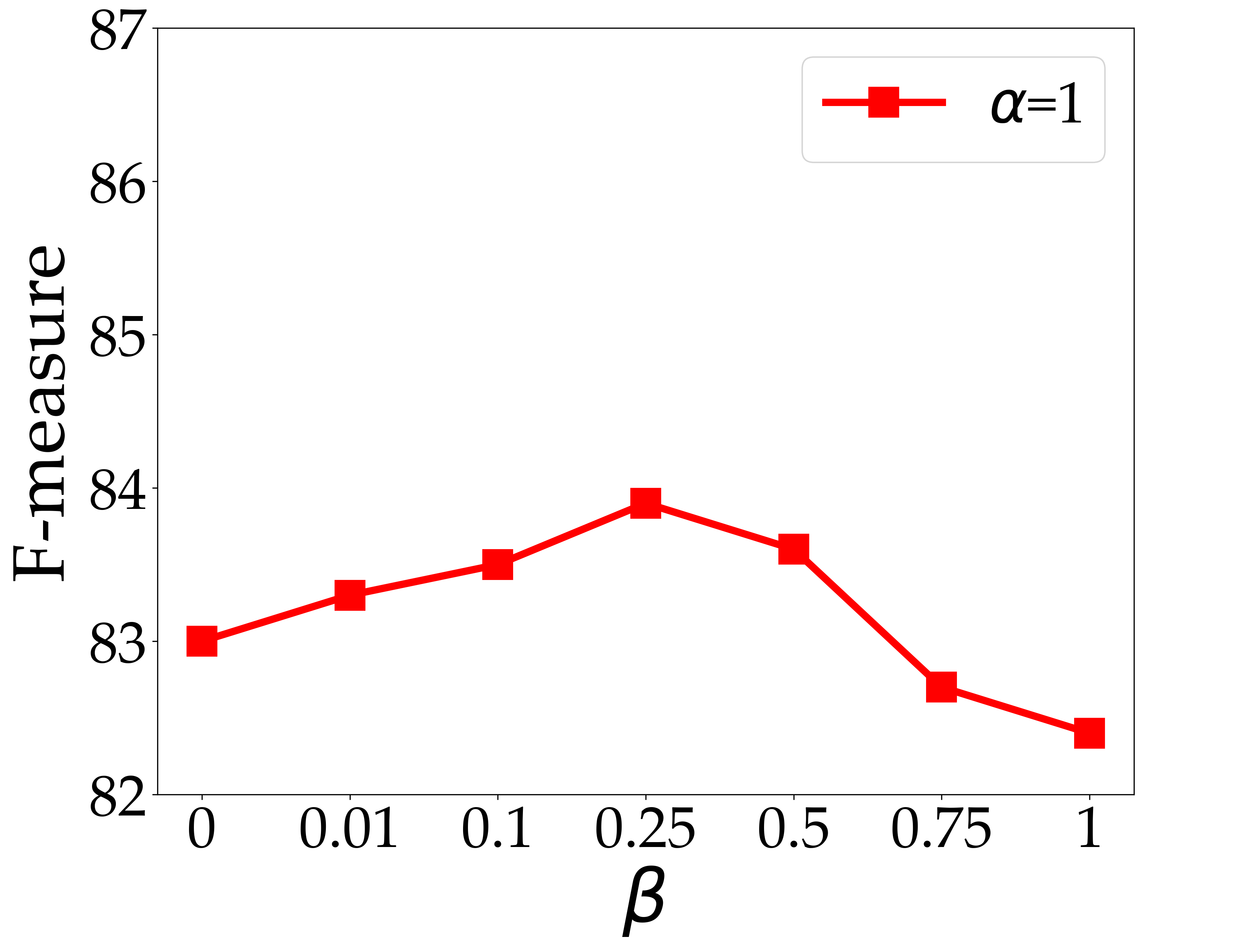}
	\end{minipage}}
	\subfigure[Impact of $\gamma$ for F-measure]{
		\begin{minipage}[b]{0.45\linewidth}
			\includegraphics[width=1\linewidth]{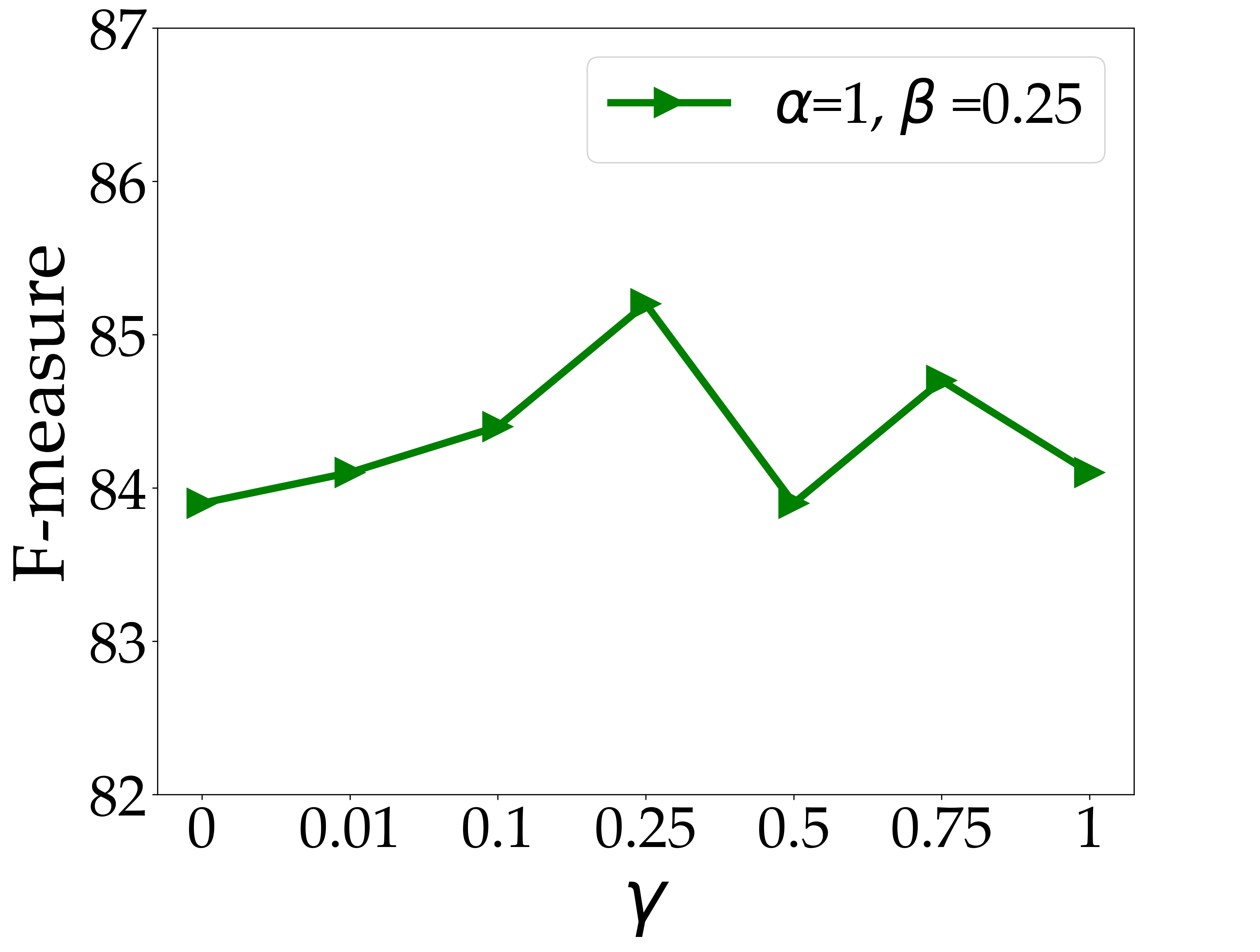}
	\end{minipage}}
	
	\subfigure[Impact of $\eta$ for F-measure]{
		\begin{minipage}[b]{0.45\linewidth}
			\includegraphics[width=1\linewidth]{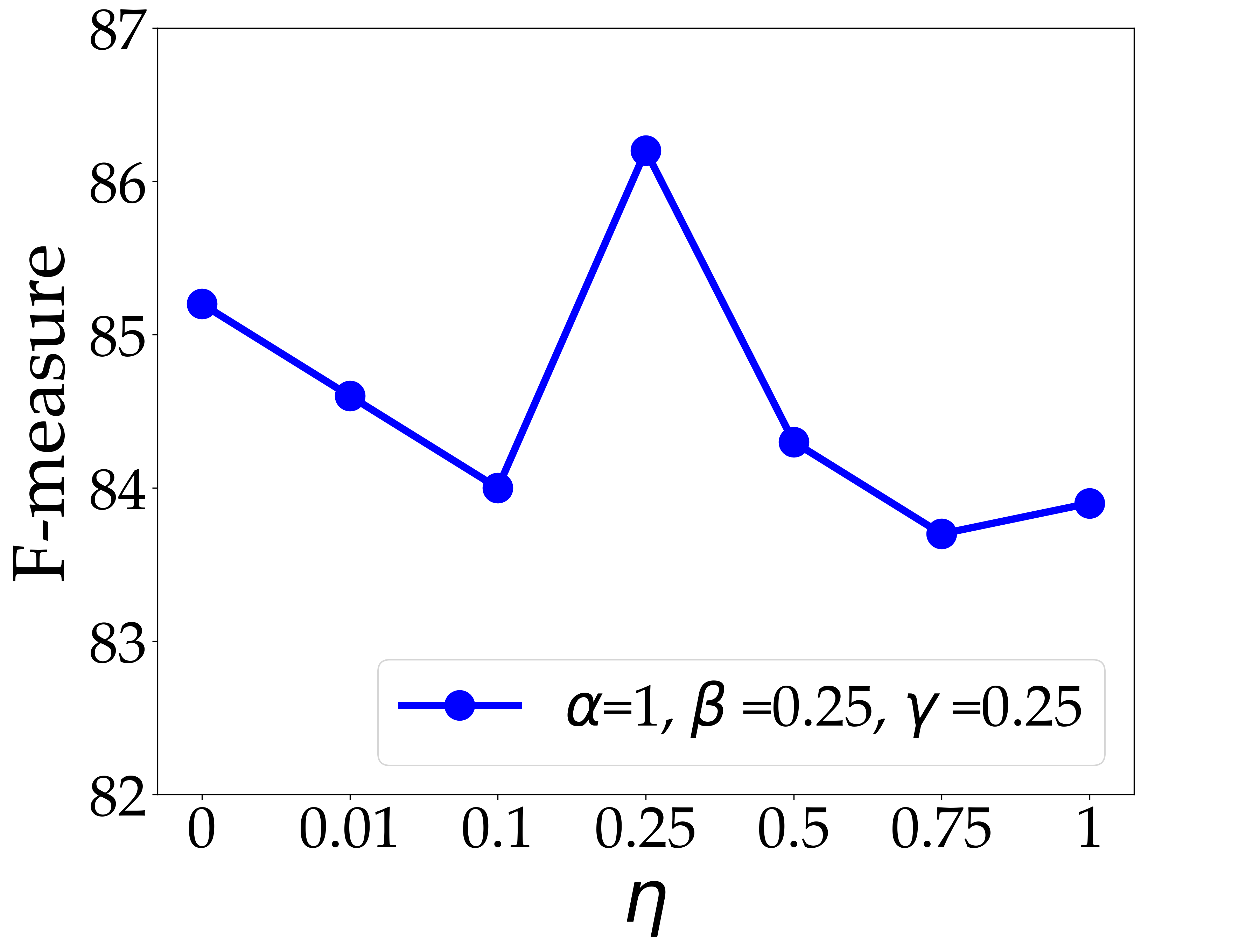}
	\end{minipage}}
	\subfigure[Impact details of $\beta$, $\gamma$, and $\eta$]{
		\begin{minipage}[b]{0.44\linewidth}
			\includegraphics[width=1\linewidth]{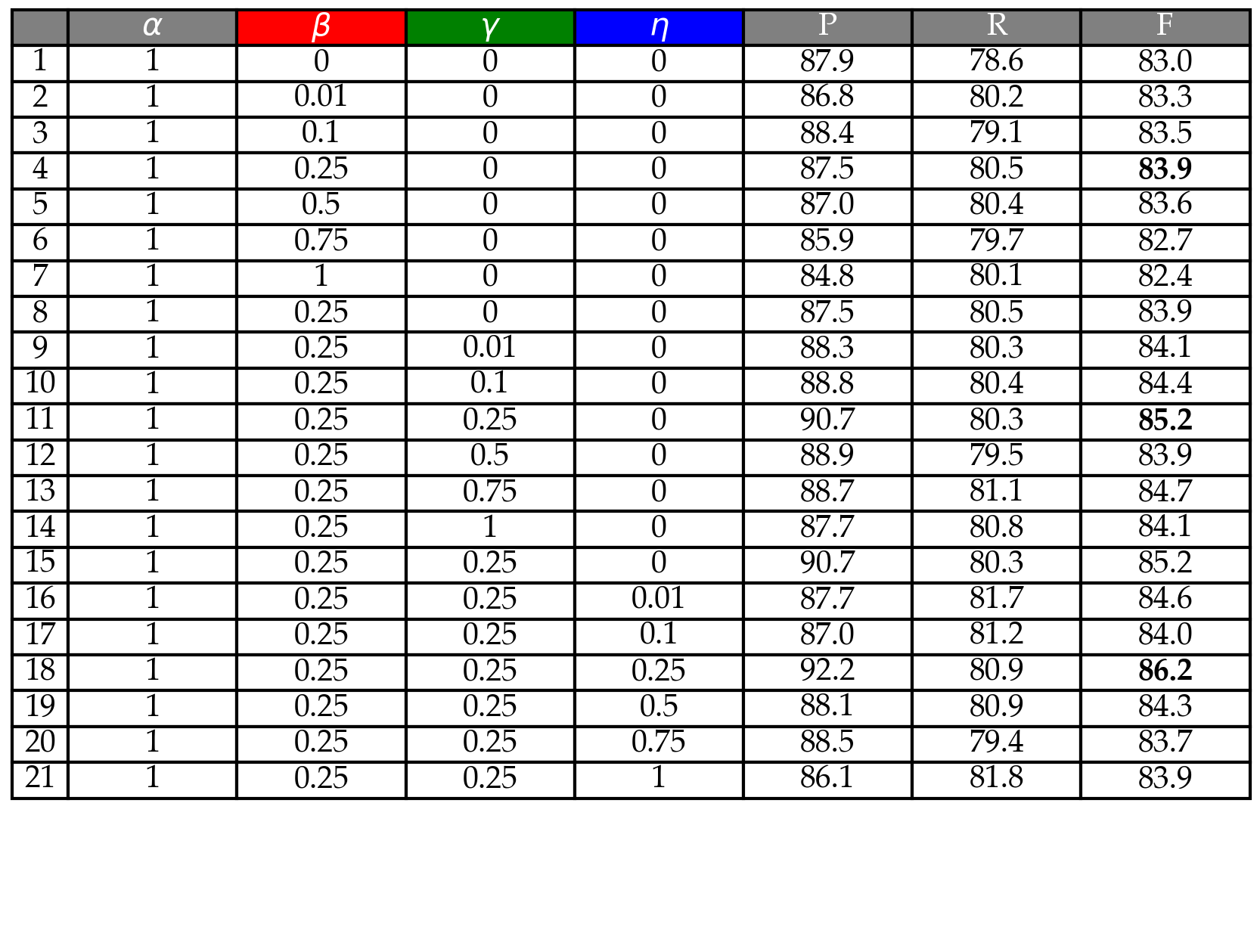}
	\end{minipage}}
	\caption{Ablation study for the impact of $\beta$, $\gamma$, and $\eta$ on performance.}
	\label{R3}
\end{figure}

\textbf{Effectiveness of Zoom In Module.} As described in Section~\ref{overall}, text contour is generated through extending shrink-mask contour outward by a specific distance, which means the accuracy of shrink-mask edge influences model performance directly. ZIM is proposed to force our method to focus on the margins, which helps ZTD to recognize shrink-mask edges precisely. Compared with baseline (Table~\ref{t1}~\#1), ZIM brings 0.9\% improvements in F-measure. Particularly, it brings no extra computational cost to the inference process, which benefits from the sharing structure between baseline and ZIM. Moreover, as we can see from the second and third columns in Fig.~\ref{R1}, ZIM encourages the baseline to perform better for the recognition of shrink-mask edges. The loss curve of the training process of baseline+ZIM is shown in Fig.~\ref{R2}, which enjoys a faster convergence speed compared to baseline.

\textbf{Effectiveness of Zoom Out Module.} As we mentioned before, ZOM is presented to avoid the phenomenon of feature defocusing to enhance the discrimination of shrink-masks from the background. It is found in Table~\ref{t1}, the \#3 model outperforms baseline 2.2\% and \#2 1.3\% in F-measure, respectively, which verifies the effectiveness of ZOM. Meanwhile, we compare the detection results of the \#3 model with baseline and baseline+ZIM in Fig.~\ref{R1}. It can be seen that ZOM helps our method to discriminate shrink-masks from some interference regions of the background effectively.

\textbf{Effectiveness of Sequential-Visual Discriminator.} Considering false-positive samples enjoy highly similar visual features with shrink-masks and are hard to recognize according to visual features only, SVD is designed to encourage ZTD to suppress them by the combination of sequential and visual features. As shown in the fourth and fifth columns in Fig.~\ref{R1}, we can find that SVD helps our method to suppress false-positive samples effectively. The experimental results in Table~\ref{t1}~\#4 also verify the effectiveness of SVD. Moreover, we pre-train our model on SynthText in this paper to keep a fair comparison environment with existing methods. As shown in Fig.~\ref{R1}~(f), pre-training our model on SythText improve the accuracy of predicted shrink-masks and brings 0.6\% F-measure (Table~\ref{t1}~\#5). Furthermore, we can see from Fig.~\ref{R2}~(a), the pre-trained model converges more faster than others.

\textbf{Impacts of Different Settings for SVD.} SVD extracts sequential features by RNN and inputs the features into FCN-based classifier to help ZTD to discriminate shrink-masks from false-positive samples (as shown in Fig.~\ref{V6}). In this section, we explore the influences of different units and the number of RNN layers for the extraction of sequential features. As shown in Fig.~\ref{t2}~(a), LSTM-based SVD brings 0.9\% and 0.4\% improvements in F-measure compared to normal RNN unit and GRU, respectively. Moreover, we can see from Fig.~\ref{t2}~(b), ZTD achieves the optimal performance when the number of RNN layers is set to 2. The above experimental results not only verify the positive effect of sequential features for the discrimination of shrink-masks but also demonstrate the prominent performance of LSTM to extract the sequential features of very long shrink-masks.

\textbf{Importances of Different Sub-losses.} As described in Section~\ref{lossfunction}, the optimization function ${\cal L}$ is composed of ${\cal L}_{sm}$, ${\cal L}_{zi}$, ${\cal L}_{zo}$, and ${\cal L}_{svd}$. $\alpha$, $\beta$, $\gamma$, and $\eta$ are the corresponding weights. In this section, we tune the value of a single weight and keep others fixed to evaluate the importance of each sub-loss. All experimental results are shown in Fig.~\ref{R3}. $\alpha$ is the weight of shrink-mask prediction header, it is set to 1 empirically. For $\beta$, $\gamma$, and $\eta$, we first analyze the importance of $\beta$. As shown in Fig.~\ref{R3}~(a), the proposed ZTD achieves the optimal performance when $\beta$ is equal to 0.25, which indicates ZIM has a certain positive effect for the prediction of the shrink-mask. Furthermore, we perform the same analysis for $\gamma$. As demonstrated in Fig.~\ref{R3}~(b), the model performance is always better than baseline+ZIM when tuning $\gamma$ in the range of 0--1, which demonstrates the effectiveness of ZOM for the distinguishment between shrink-masks and the background. Moreover, we test $\eta$ by the same experiment. As shown in Fig.~\ref{R3}~(c), ZTD achieves the best performance when $\eta$ is set to 0.25 and the performance fluctuates when $\eta$ is close to 0.1 and 0.75. In Fig.~\ref{R3}~(d), the impact details of $\beta$, $\gamma$, and $\eta$ on model performance are described, which helps to understand the impartances of different sub-losses intuitively.

\subsection{Comparison with State-of-the-Art Methods}
\label{comparison}
To verify the superior performance of ZTD, we compare it with the existing competitors on multiple representative public benchmarks (such as MSRA-TD500, Total-Text, CTW1500, and ICDAR2015) in this section. Considering existing text detection methods can be categorized into accuracy prior and comprehensive performance prior methods roughly (as mentioned in Section~\ref{sec2}), we analyze the advantages of ZTD over them respectively in the following experiments.

\begin{table}[]
	\renewcommand{\arraystretch}{1.1}
	\setlength{\tabcolsep}{1.6mm}
	\caption{Performance comparison on MSRA-TD500 Dataset.}
	\centering
	\begin{tabular}{lcccc}
		\toprule
		Methods                                               &   P  &   R  &   F  &  FPS \\ \midrule
		$Accuracy~Prior$  &  &  &  &  \\ 
		PixelLink~\cite{DBLP:conf/aaai/DengLLC18} (AAAI~2018) & 83.0 & 73.2 & 77.8 &  -   \\ 
		RRD~\cite{liao2018rotation} (CVPR~2018)               & 87.0 & 73.0 & 79.0 &  10  \\  
		CRAFT~\cite{baek2019character} (CVPR~2018)            & 88.2 & 78.2 & 82.9 & 8.6  \\ 
		SAE~\cite{tian2019learning} (CVPR~2019)               & 84.2 & 81.7 & 82.9 & -    \\
		TexrField~\cite{xu2019textfield} (TIP~2019)           & 87.4 & 75.9 & 81.3 &  -   \\ 
		OPMP~\cite{zhang2020opmp} (TMM~2020)                  & 86.0 & 83.4 & 84.7 & 1.6  \\ 
		SAVTD~\cite{DBLP:conf/cvpr/FengYZL21} (CVPR~2021)     & 89.2 & 81.5 & 85.2 & -    \\ 
		GV~\cite{xu2020gliding} (TPAMI~2021)                  & 88.8 & 84.3 & 86.5 & 15.0 \\ \midrule
		$Comprehensive~Performance~Prior$  &  &  &  &  \\ 
		DB~\cite{liao2020real} (AAAI~2020)                    & 90.4 & 76.3 & 82.8 & 62.0 \\  
		PAN~\cite{wang2019efficient} (ICCV~2019)              & 84.4 & 83.8 & 84.1 & 30.2 \\
		PAN++~\cite{wang2021pan++} (TPAMI~2021)               & 85.3 & 84.0 & 84.7 & 32.5 \\ \midrule
		\rowcolor{lightgray!80}
		ZTD-512 (Ours)                                        & 90.5 & 82.1 & 86.1 & 97.4\\ 
		\rowcolor{lightgray!80}
		ZTD-640 (Ours)                                        & 91.5 & 81.6 & 86.3 & 72.7 \\ 
		\rowcolor{lightgray!80}
		ZTD-736 (Ours)                                        & 91.6 & 82.4 & 86.8 & 59.2 \\ \bottomrule
	\end{tabular}
	\label{t3}
	\vspace{-3mm}
\end{table}

\begin{figure*}
	\begin{center}
		\includegraphics[width=0.9\textwidth]{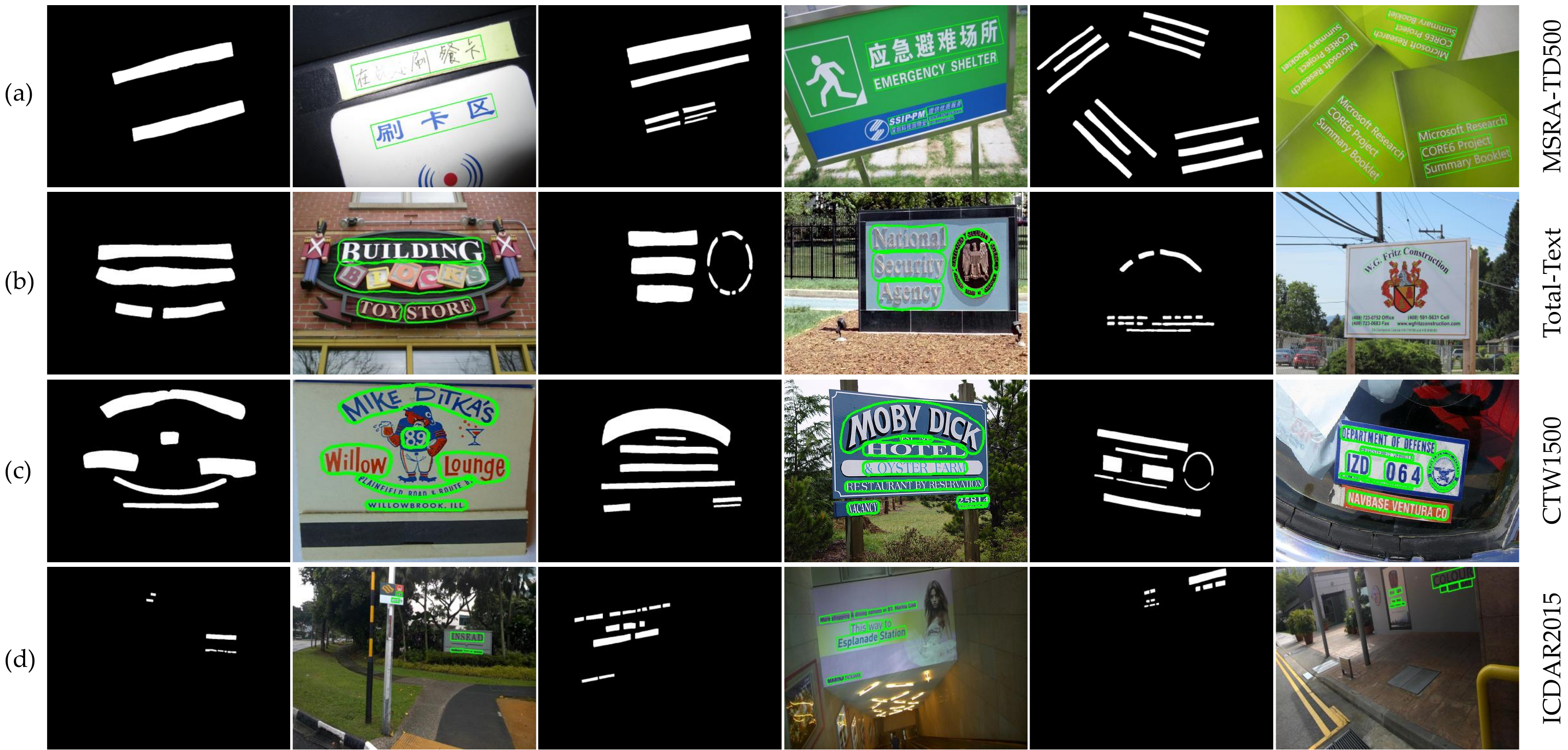}
	\end{center}
	\vspace{-5mm}
	\caption{Visualization of some qualitative detection results of ZTD on MSRA-TD500, Total-Text, CTW1500, and ICDAR2015 datasets. Binary masks are the predicted shrink-masks and RGB images show the rebuilt text contours based on the predicted shrink-masks.}
	\label{R4}
\end{figure*}

\textbf{Evaluation on MSRA-TD500.} We evaluate the performance of ZTD for detecting multi-language long straight text instances on MSRA-TD500 dataset. The experimental results are shown in Table~\ref{t3}. It is found that our method outperforms existing state-of-the-art (SOTA) approaches in both detection accuracy and speed. Specifically, for GV~\cite{xu2020gliding}, the best accuracy prior method, ZTD-736 outperforms it by 0.3\% in F-measure. It is because the proposed ZIM, ZOM, and SVD enhance the model's ability to recognize shrink-masks. Meanwhile, benefiting from the lightweight CNN model and simple post-processing, our method runs 4 times faster than it. Furthermore, the comprehensive performance of ZTD-736 outperforms PAN~\cite{wang2019efficient}, PAN++~\cite{DBLP:conf/aaai/DengLLC18} a lot. Though DB~\cite{liao2020real} achieves 62.0 FPS in detection speed, ZTD-512 is 35.4 FPS faster than it. Some qualitative results are shown in Fig.~\ref{R4}~(a). The experiments on MSTA-TD500 demonstrate the effectiveness of ZTD for detecting long text instances, even they are multilingual. 

\begin{table}[]
	\centering
	\renewcommand{\arraystretch}{1.1}
	\setlength{\tabcolsep}{1.6mm}
	\caption{Performance comparison on Total-Text Dataset.}
	\begin{tabular}{lcccc}
		\toprule
		Methods                                               &   P  &   R  &   F  &  FPS \\ \midrule
		$Accuracy~Prior$  &  &  &  &  \\ 
		TextSnake~\cite{long2018textsnake} (ECCV~2018)        & 82.7 & 74.5 & 78.4 &  -   \\ 
		TextDragon~\cite{feng2019textdragon} (ICCV~2019)      & 85.6 & 75.7 & 80.3 &  -   \\ 
		TextField~\cite{xu2019textfield} (TIP~2019)           & 81.2 & 79.9 & 80.6 &  -   \\ 
		Boundary~\cite{wang2020all} (AAAI~2020)             & 85.2 & 83.5 & 84.3 &  -   \\
		ContourNet~\cite{wang2020contournet} (CVPR~2020)      & 86.9 & 83.9 & 85.4 & 3.8  \\
		DRRG~\cite{DBLP:conf/cvpr/ZhangZHLYWY20} (CVPR~2020)  & 86.5 & 84.9 & 85.7 &  -   \\ 
		FCENet~\cite{zhu2021fourier} (CVPR~2021)              & 87.4 & 79.8 & 83.4 &  -   \\
		ReLaText~\cite{ma2021relatext} (PR~2021)              & 84.8 & 83.1 & 84.0 &  - \\ 
		MaskTextSpotter~\cite{8812908} (TPAMI~2021)           & 88.3 & 82.4 & 85.2 & - \\   
		\midrule
		$Comprehensive~Performance~Prior$  &  &  &  &  \\ 
		DB~\cite{liao2020real} (AAAI~2020)                    & 88.3 & 77.9 & 82.8 & 50.0 \\  
		PAN~\cite{wang2019efficient} (ICCV~2019)              & 89.3 & 81.0 & 85.0 & 39.6 \\
		PAN++~\cite{wang2021pan++} (TPAMI~2021)               & 89.9 & 81.0 & 85.3 & 38.3 \\
		KPN~\cite{zhang2022kernel} (TNNLS~2022)               & 88.0 & 82.3 & 85.1 & 22.7 \\ \midrule
		\rowcolor{lightgray!80}
		ZTD-512 (Ours)                                        & 90.5 & 80.6 & 85.3 & 93.2 \\ 
		\rowcolor{lightgray!80}
		ZTD-640 (Ours)                                        & 90.1 & 82.3 & 86.0 & 75.2 \\ \bottomrule
	\end{tabular}
	\label{t4}
	\vspace{-3mm}
\end{table}

\textbf{Evaluation on Total-Text.} To verify the robustness of ZTD to detect word-level irregular-shaped texts, we evaluate it on Total-Text benchmark. The same as the experimental conclusion on MSRA-TD500, our method is superior to others in both detection accuracy and speed. As shown in Table~\ref{t4}, for accuracy prior methods, MaskTextSpotter~\cite{8812908}, ContourNet~\cite{wang2020contournet}, and DRRG~\cite{DBLP:conf/cvpr/ZhangZHLYWY20} achieve 85.2\%, 85.4\%, and 85.7\% in F-measure, respectively. For comprehensive performance prior approaches, PAN~\cite{wang2019efficient} and PAN++~\cite{DBLP:conf/aaai/DengLLC18} enjoy comparable detection accuracy with accuracy prior methods. DB~\cite{liao2020real} performs better in detection speed. 
Compared with the above methods, ZTD can achieve 86.0\% in F-measure and 75.2 FPS. Since many texts are close to each other in Total-Text, existing methods are hard to separate them efficiently. Benefiting from ZIM, our method enjoys a strong ability to recognize shrink-mask edges, which helps ZTD to avoid text adhesion problem. We further display some detection results in Fig.~\ref{R4}~(b). It is found that adhesive texts are separated successfully. 

\begin{table}[]
	\renewcommand{\arraystretch}{1.1}
	\setlength{\tabcolsep}{1.6mm}
	\caption{Performance comparison on CTW1500 Dataset.}
	\centering
	\begin{tabular}{lcccc}
		\toprule
		Methods                                               &   P  &   R  &   F  &  FPS \\ \midrule
		$Accuracy~Prior$  &  &  &  &  \\ 
		CRAFT~\cite{baek2019character} (CVPR~2018)            & 86.0 & 81.1 & 83.5 &  -   \\ 
		LOMO~\cite{zhang2019look} (CVPR~2019)                 & 85.7 & 76.5 & 80.8 &  -   \\ 
		TexrField~\cite{xu2019textfield} (TIP~2019)           & 83.0 & 79.8 & 81.4 &  -   \\
		OPMP~\cite{zhang2020opmp} (TMM~2020)                  & 85.1 & 80.8 & 82.9 &  1.4   \\
		TextRay~\cite{wang2020textray} (ACMMM~2020)           & 82.8 & 80.4 & 81.6 &  -   \\ 
		DRRG~\cite{DBLP:conf/cvpr/ZhangZHLYWY20} (CVPR~2020)  & 85.9 & 83.0 & 84.5 &  -   \\
		FCENet~\cite{zhu2021fourier} (CVPR~2021)              & 85.7 & 80.7 & 83.1 &  -   \\
		ReLaText~\cite{ma2021relatext} (PR~2021)              & 86.0 & 83.3 & 84.8 & 10.6 \\  
		\midrule
		$Comprehensive~Performance~Prior$  &  &  &  &  \\ 
		DB~\cite{liao2020real} (AAAI~2020)                    & 84.8 & 77.5 & 81.0 & 55.0 \\  
		PAN~\cite{wang2019efficient} (ICCV~2019)              & 86.4 & 81.2 & 83.7 & 39.8 \\
		PAN++~\cite{wang2021pan++} (TPAMI~2021)    & 87.1 & 81.1 & 84.0 & 36.0 \\ 
		KPN~\cite{zhang2022kernel} (TNNLS~2022)    & 84.0 & 82.9 & 83.4 & 24.3 \\ \midrule
		\rowcolor{lightgray!80}
		ZTD-640 (Ours)                                        & 88.4 & 80.2 & 84.1 & 76.9 \\ \bottomrule
	\end{tabular}
	\label{t5}
	\vspace{-3mm}
\end{table}

\textbf{Evaluation on CTW1500.} Experiments on CTW1500 show the effectiveness of ZTD for detecting line-level arbitrary-shaped texts. All experimental results are shown in Table~\ref{t5}. Our method runs 76.9 FPS and is faster than other methods at least by 21.9 FPS. The outstanding detection speed benefits from the efficient text representation method and lightweight CNN model. Specifically, PAN~\cite{wang2019efficient} and PAN++~\cite{DBLP:conf/aaai/DengLLC18} reconstruct text contours through pixel-wise extension strategy. Unlike these comprehensive performance prior methods, ZTD adopts an object-wise extension strategy. Moreover, our method optimizes the CNN model to design a lightweight and efficient network, which significantly gains our approach in detection speed. For accuracy prior methods, though the detection accuracy of ZTD is not as well as some methods (such as DRRG~\cite{DBLP:conf/cvpr/ZhangZHLYWY20} and ReLaText~\cite{ma2021relatext}), the proposed approach has at least 7 times faster speed than them. As shown in Fig.~\ref{R4}~(c), the above experiments and the visualization of detection results demonstrate the effectiveness of ZTD to recognize long irregular-shaped text instances.

\begin{table}[]
	\renewcommand{\arraystretch}{1.1}
	\setlength{\tabcolsep}{1.6mm}
	\caption{Performance comparison on ICDAR2015 Dataset.}
	\centering
	\begin{tabular}{lcccc}
		\toprule
		Methods                                               &   P  &   R  &   F  &  FPS \\ \midrule
		$Accuracy~Prior$  &  &  &  &  \\  
		TextSnake~\cite{long2018textsnake} (ECCV~2018)        & 84.9 & 80.4 & 82.6 &  - \\ 
		CornerNet~\cite{law2018cornernet} (ECCV~2018)         & 89.5 & 79.7 & 84.3 &  1.0 \\
		PSE~\cite{wang2019shape} (CVPR~2019)                  & 86.9 & 84.5 & 85.7 &  1.6 \\  
		SAE~\cite{tian2019learning} (CVPR~2019)               & 85.1 & 84.5 & 84.8 &  - \\ 
		Boundary~\cite{wang2020all} (AAAI~2020)               & 88.1 & 82.2 & 85.0 &  - \\
		FCENet~\cite{zhu2021fourier} (CVPR~2021)              & 85.1 & 84.2 & 84.6 & -    \\
		TEETS~\cite{wang2021towards} (TPAMI~2021)             &  -   &  -   & 85.0 & -    \\
		MaskTextSpotter~\cite{8812908} (TPAMI~2021)           & 85.8 & 81.2 & 83.4 & 4.8  \\
		PolarMask++~\cite{xie2021polarmask++} (TPAMI~2021)    & 87.3 & 83.5 & 85.4 & 10.0 \\
		\midrule
		$Comprehensive~Performance~Prior$  &  &  &  &  \\ 
		DB~\cite{liao2020real} (AAAI~2020)                    & 86.8 & 78.4 & 82.3 & 48.0 \\  
		PAN~\cite{wang2019efficient} (ICCV~2019)              & 84.0 & 81.9 & 82.9 & 26.1 \\
		PAN++~\cite{wang2021pan++} (TPAMI~2021)               & 85.9 & 80.4 & 83.1 & 28.2 \\ \midrule
		\rowcolor{lightgray!80}
		ZTD-736 (Ours)                                        & 87.5 & 79.0 & 83.0 & 48.3\\
		\bottomrule
	\end{tabular}
	\label{t6}
	\vspace{-3mm}
\end{table}

\textbf{Evaluation on ICDAR2015.} To verify the robustness of ZTD to detect multi-oriented text instances from the complicated background, we compare ZTD with existing text detection methods on ICDAR 2015 benchmark. As exhibited in Table~\ref{t6}, our method achieves 83.0\% F-measure with 48.3 FPS, which outperforms DB~\cite{liao2020real} and PAN~\cite{wang2019efficient} in both detection accuracy and speed. Moreover, ZTD can run 2 times faster than PAN++~\cite{DBLP:conf/aaai/DengLLC18} and achieves comparable detection accuracy to it. Compared with the accuracy prior methods, the proposed detector keeps considerable superiority in detection speed and accomplishes the best balance between detection accuracy and speed. The superior comprehensive performance brings great potential for a wide range of applications. The results in Table~\ref{t6} and Fig.~\ref{R4}~(d) demonstrate our method can recognize the texts with various scales and multi-orientations from the complex background effectively.

\begin{table}[]
	\renewcommand{\arraystretch}{1.1}
	\setlength{\tabcolsep}{1.69mm}
	\caption{Cross-dataset evaluations on word-level and line-level datasets.}
	\centering
	\begin{tabular}{cccccc}
		\toprule
		Type                        & Traning dataset & Test dataset &   P  &   R  &   F  \\ \midrule
		\multirow{2}{*}{word-level} & ICDAR2015       & Total-Text      & 78.5 & 64.1 & 70.6 \\ 
		& Total-Text      & ICDAR2015       & 79.8 & 69.3 & 74.2 \\ 
		\multirow{2}{*}{Line-level} & MSRA-TD500      & CTW1500         & 84.1 & 73.4 & 78.4 \\ 
		& CTW1500         & MSRA-TD500      & 86.8 & 77.9 & 82.1 \\ \bottomrule
	\end{tabular}
	\label{t7}
\end{table}

\subsection{Cross Dataset Text Detection}
We conduct multiple comparison experiments in Section~\ref{comparison} and show the superior performance in both detection accuracy and speed of our method. To further verify the generalization performance of ZTD, we further evaluate it through a series of cross-train-test experiments. Specifically, considering ICDAR2015 and Total-Text are word-level datasets, MSRA-TD500 and CTW1500 are line-level benchmarks, we design two sets of experiments on word-level and line-level datasets, respectively. At first, our method is trained on the training images of ICDAR2015 and Total-Text. Then, we evaluate ZTD on the testing images of Total-Text and ICDAR2015. As we can see from Table~\ref{t7}, ZTD achieves 70.6\% and 74.2\% in F-measure, respectively. For line-level datasets, the same cross-train-test experiments are conducted. Particularly, our method achieves 82.1\% in F-measure when it is trained on CTW1500 and tested on MSRA-TD500, which surpasses many methods (e.g., PixelLink~\cite{DBLP:conf/aaai/DengLLC18}, RRD~\cite{liao2018rotation}, and TextField~\cite{xu2019textfield}) that is trained on MSRA-TD500, which shows the effectiveness of our method for long text detection and the generalization performance in different scenes.

\begin{table}[]
	\renewcommand{\arraystretch}{1.1}
	\setlength{\tabcolsep}{0.9mm}
	\caption{Time consumption of ZTD on four public benchmarks. Different stages of ZTD include Backbone, Zoom (ZIM and ZOM), Head (shrink-mask prediction header), and Post (contour extension process). `Image size' denotes the size of the image short side.}
	\centering
	\begin{tabular}{cccccccc}
		\toprule
		\multirow{2}{*}{Datasets} & \multirow{2}{*}{Image size} & \multicolumn{4}{c}{Time consumption (ms)}                     & \multirow{2}{*}{FPS} & \multirow{2}{*}{F} \\ \cline{3-6}
		&                         & \multicolumn{1}{c}{Backbone} & \multicolumn{1}{c}{Zoom} & \multicolumn{1}{c}{Head} & Post &                         &                            \\ \midrule
		MSRA-TD500                & 736                         & \multicolumn{1}{c}{8.0}      & \multicolumn{1}{c}{4.2}  & \multicolumn{1}{c}{3.3}   & 1.4  & 59.2                 & 86.8                       \\ 
		Total-Text                & 640                         & \multicolumn{1}{c}{6.2}      & \multicolumn{1}{c}{3.1}  & \multicolumn{1}{c}{2.6}   & 1.4  & 75.2                 & 86.0                       \\
		CTW1500                   & 640                         & \multicolumn{1}{c}{6.0}      & \multicolumn{1}{c}{3.1}  & \multicolumn{1}{c}{2.5}   & 1.4  & 76.9                 & 84.1                       \\ 
		ICDAR2015                 & 736                         & \multicolumn{1}{c}{9.9}      & \multicolumn{1}{c}{5.1}  & \multicolumn{1}{c}{4.1}   & 1.6  & 48.3                 & 83.0                       \\ \bottomrule
	\end{tabular}
	\label{t8}
\end{table}

\subsection{Speed Analysis} 
The above experiments demonstrate the outstanding comprehensive performance of our method. Especially in terms of detection speed, the proposed ZTD enjoys an obvious advantage compared to previous algorithms. To verify the high efficiency of the designed framework, we analyze the time consumption details of ZTD's different stages in this section. The experimental details as described in Table~\ref{t8}. To keep a fair comparison environment, we resize the short sides of images as 736, 640, 640, and 736 for MSRA-TD500, Total-Text, CTW1500, and ICDAR2015, respectively. It is found that 'Backbone` takes about half of the total time. It is mainly because 'Backbone` is composed of plenty of convolution layers. Unlike 'Backbone`, 'Zoom` and 'Head` are composed of fewer convolution layers. However, as a decoder structure, 'Head' needs to upsample feature maps to image size, which increases the time consumption. Therefore, though the layers of 'Zoom` and 'Head` are less than 'Backbone`, they almost consume the same computational cost as 'Backbone`. The 'Post` denotes the contour extension process (shown in Fig.~\ref{V3}). Since it is an object-wise operation, the time consumption is much less than the above stages and does not influenced by the image size (as the comparison between different datasets in Table~\ref{t8}). The lightweight CNN model and object-wise contour extension process bring significant improvements for our method in detection speed, and the experimental results demonstrate this conclusion.

\subsection{Failuer Cases} 
\begin{figure}
	\begin{center}
		\includegraphics[width=0.48\textwidth]{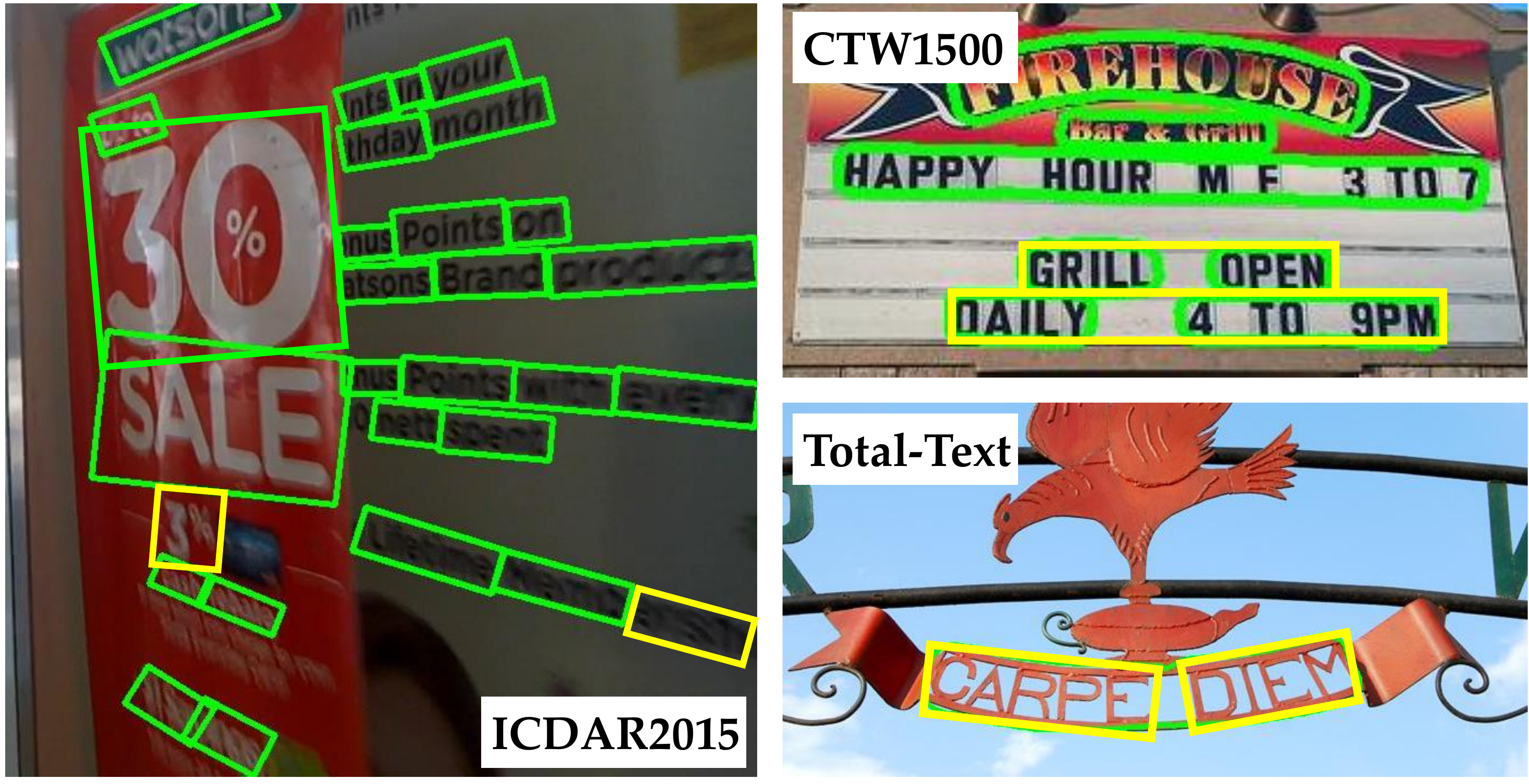}
	\end{center}
	\vspace{-5mm}
	\caption{Challenging samples from ICDAR2015, CTW1500, and Total-Text datasets. The green bounding boxes are the detection results from our method. The yellow ones are labels.}
	\label{R5}
\end{figure}

We have verified the superiority of the proposed ZTD in both detection accuracy and speed on multiple public benchmarks before. To further analyze the limitation of the proposed detector, we show some incorrect detection results. As demonstrated in Fig.~\ref{R5}, there are three challenging samples from ICDAR2015, CTW1500, and Total-Text datasets, respectively. For the sample from ICDAR2015, two text instances are missed detection, where blurred, and low color contrast are the failure reasons. For line-level (CTW1500) and word-level (Total-Text) datasets, half detection and overdetection are the current main problems, respectively. The above issues make there is still much room to improve the proposed method.

\section{Conclusion}
\label{sec5}
In this paper, we propose an efficient text detector inspired by the zoom process of the camera, termed as Zoom Text Detector (ZTD). By simulating the zooming out process of the camera, the detector can extract strong expressive semantic features from coarse layers, which enhances ZTD's ability to discriminate shrink-masks from the background significantly. Moreover, simulating the zooming in process of the camera encourages our method to focus on the margins, which helps to recognize shrink-mask edges accurately and avoid many problems (e.g., text adhesion and missed detection). Additionally, sequential features are extracted and combined with visual features to facilitate the presented approach to suppress false-positive samples effectively, which further improves the reliability of predicted shrink-masks. Extensive experiments show the effectiveness of ZOM, ZIM, and SVD. Comparisons on the multiple benchmarks demonstrate the superior comprehensive performance in both detection accuracy and speed of ZTD, which shows the great potential for a wide range of applications.

\bibliographystyle{IEEEtran}
\bibliography{egbib}

\vfill

\end{document}